\documentclass[a4paper,11pt]{article}
\usepackage[left=2.5cm,right=2.5cm,top=3cm,bottom=3cm]{geometry}
\usepackage[utf8]{inputenc}

\usepackage{xcolor}
\definecolor{blendedblue}{rgb}{0.2, 0.2, 0.6}
\definecolor{blendedred}{rgb}{0.8, 0.2, 0.2}
\usepackage[bookmarks=true,
            breaklinks=true,
            pdfborder={0 0 0},
            citecolor=blendedblue,
            colorlinks=true,
            linkcolor=blendedblue,
            urlcolor=blendedblue,
            citecolor=blendedblue,
            linktocpage=false,
            hyperindex=true,
            linkbordercolor=white]{hyperref}

\usepackage[american]{babel}
\usepackage{lineno}
\usepackage{csquotes}
\usepackage{sansmath}
\usepackage{amsmath, amssymb, amsthm, mathtools}
\usepackage{nicefrac}
\usepackage[ttscale=.875]{libertine}
\usepackage[scaled=0.96]{zi4}
\usepackage{adjustbox}
\usepackage{parskip}
\usepackage{graphicx}
\graphicspath{{./img/}}

\usepackage[numbers]{natbib}

\usepackage[labelsep=period]{caption}

\usepackage{enumitem} 
\usepackage{caption}
\usepackage{subcaption}
\usepackage{algorithm}
\usepackage{algorithmic}
\usepackage{amsmath}
\usepackage{amsfonts}
\usepackage{amssymb}

\newlength\myindent
\newcommand\bindent{%
  \begingroup
  \setlength{\itemindent}{\myindent}
  \addtolength{\algorithmicindent}{\myindent}
}
\newcommand\eindent{\endgroup}

\begin{document}

\title{\textbf{Knowledge extraction from the learning of sequences\\
               in a long short term  memory (LSTM) architecture}}
\author{\textbf{Ikram Chraibi Kaadoud}$^{1,2}$ \and
        \textbf{Nicolas Rougier}$^{1,2,3}$ \and 
        \textbf{Frédéric Alexandre}$^{1,2,3,*}$\\
$^1$ Inria Bordeaux Sud-Ouest, Talence, France\\
$^2$ LaBRI, UMR 5800, CNRS, Université de Bordeaux, Talence, France\\
$^3$ Institut des Maladies Neurodégénératives, UMR 5293, CNRS,\\ Université de Bordeaux, Bordeaux, France\\
* Corresponding author: Frederic.Alexandre@inria.fr}

\maketitle

\begin{abstract}

We introduce a general method to extract knowledge from a recurrent neural network (Long Short Term Memory) that has learnt to detect if a given input sequence is valid or not, according to an unknown generative automaton. Based on the clustering of the hidden states, we explain how to build and validate an automaton that corresponds to the underlying (unknown) automaton, and allows to predict if a given sequence is valid or not. The method is illustrated on artificial grammars (Reber’s grammar variations) as well as on a real use-case whose underlying grammar is unknown.


\end{abstract}


\setcounter{tocdepth}{2}
\tableofcontents
\clearpage

\section{Introduction}
\label{intro}



Efficient numerical models are exploited as standard tools to process data in many domains of the socio-economic world. The possibility to extract knowledge from this kind of models becomes an increasingly important demand, as it is frequently discussed for ANNs (Artificial Neural Networks). It is a valuable asset when one wants to compare such models with more traditional and explicit algorithmic approaches. It is even more fundamental, not to say mandatory, in critical domains where the functioning of an automatic decision-making system must be assessed by humans, in order to check that it relies on valid cues. Transparency is often evoked to justify this requirement \citep{lipton2016}, advocating that the end-user of the model should have a level of understanding comparable to the expert, designer of the model and owner of the hidden knowledge. 
This is also true when the underlying knowledge is not known a priori and when the assessment is not a simple verification but rather corresponds to the discovery of a hidden knowledge that might be confirmed a posteriori. 

This situation can be compared, in human cognition, to the distinction between explicit and implicit memory \citep{reber_neural_2013}. In the explicit case, the knowledge is conscious and human decisions can be explained and expressed in a declarative way (you can for example explain that you have ranked students by comparing their marks following explicit rules). In the implicit case, you can acquire skills without being conscious of what you have learned and without being able to explain it. Consider for example typing on a keyboard a series of symbols where repeated sequences are hidden \citep{pascual1995procedural}. You improve your motor skills of key reaching without being conscious of the existence of regularities. 
Language is another typical example of implicit learning acquired by children through practice (repetition and imitation), not through explicit learning of valid rules \citep{oudeyer2009auto}. Language has been studied as an implicit learning phenomenon for different tasks and populations (adults and children)\citep{reber1967implicit, ServanSchreiber1988encoding, cleeremans1991learning, cleeremans1993mechanisms, gasparini2004implicit, gombert2006epi, nadeau2011connaissances}.

Implicit learning is pervasive in the socio-economic world because experts in a domain tend to acquire their expertise through practice and are, most of the time, unable to verbalize their knowledge (or only partly). Such expertise can also be acquired by models like ANNs by learning from examples produced by experts in orer to reproduce the corresponding skill.
In this case, the importance for knowledge extraction is high because this knowledge is hidden to all the actors of the process and might bring valuable information. As reviewed below, many approaches for knowledge extraction usin ANNs have been proposed.

In this paper, we will consider the process of knowledge extraction from implicit learning of temporal behaviour. This means that the numerical models to consider will have to be able to process sequences (as it is the case with Recurrent Neural Networks (RNNs) for example) and the data to be manipulated will be consequently corpuses of sequences. 
Preliminary attempts for extracting knowledge from simple RNN models have already been proposed. In the present paper, we propose to consider more complex and powerful RNNs, namely LSTM (Long Short Term Memory) networks \citep{hochreiter1997long}.
Before diving into the methodology, we will first discuss further some important aspects of the scientific context of  knowledge extraction in machine learning (section \ref{sec:interpretability}), introduce contexts that are classically considered in implicit learning of sequences (section \ref{UseCases}) and summarize the current state of the art (section \ref{existing}) of knowledge extraction in this case. 

\subsection{Interpretability in machine learning}
 \label{sec:interpretability}

The principles of learning algorithms associated to multi-layered supervised neuronal architectures are known for more than thirty years \citep{rumel86}. These architectures have recently gained a renewed interest mainly because larger corpuses and the corresponding computing power are now available. This led to the development of innovative architectures such as for example convolutional networks in the domain of image classification \citep{Goodfellow2016Convolutional} or Long Short Term Memory (LSTM) models for temporal sequence processing \citep{hochreiter1997long}. These latter models have outperformed simple recurrent networks (SRN) in sequential data processing and are able to deal with complex structured sequences.

However, and in spite of these major improvements, a major weakness remains for these multi-layered architectures: They are considered as black boxes and their explainability, or said differently their capacity to explain how they build their decision in a way intelligible for humans, is very weak or even non existent. 
Several notions have been proposed to describe processes of knowledge extraction from these black boxes \citep{DBLP:journals/corr/abs-1802-01933, ayache:hal-01888514}. Whereas explainability is the ultimate step where the functioning of the black box is fully understood and can be transferred to humans in understandable terms, interpretability refers to the capacity of breaking down all the inner mechanisms of the black box (without necessarily understanding them). This might be useful for an expert but not always for the layman. 
Interpretability is considered "global" when its purpose is to explicit the whole logic of the model (for example approximating the non linear black box by a simple linear model), whereas it is defined "local" when the process focuses on understanding specific reasons for a decision on a datum (for example extracting a surface of separation between categories). These examples also indicate that interpretability can be performed by an approximation process and can give an intuition rather than the full knowledge \citep{DBLP:journals/corr/abs-1802-01933}. These concepts (i.e explainability and interpretability) are often mixed in the literature, even if they are distinct since an interpretable model is a requirement for having an explanation which also depends, among other things, on the intended purpose, the decision-maker, and the context of the situation. In this paper, we will remain on the technical side and focus on the concept of interpretability and its dimensions.

In this specific field, \cite{remm02} have defined two approaches for extracting knowledge from neural networks, seen as black boxes : the pedagogical method when rules can be extracted directly from the operating network
, and the decompositional method, when elements of the rule are extracted from the hidden layers 
before being recombined globally.
As for RNNs, in the pedagogical case, some methods have tried to evaluate the equivalence between hypotheses and the RNN using queries and counter-examples to query the model in order to get pairs of (input, output) in order to model the global behaviour of the network \citep{DBLP:journals/corr/abs-1711-09576} without considering what happens inside. 
In the decompositional case, and specifically in the post-hoc interpretability, that we will consider and describe more precisely, knowledge is extracted a posteriori from the hidden layers of the network \citep{lipton2016}. 
Indeed, it is admitted that during the learning phase, the network will extract automatically simple hints or features (knowledge of features) or privileged relations expressing simple causal relations between the input and the output spaces (knowledge of rules) that are relevant for the targeted task. This knowledge representing the implicit representation of the network is encoded into its latent space \citep{abu90, bengio2013representation} which is "hidden-layer-dependant". 
So by extracting and analysing the activity of the hidden layers, during the test phase (once the learning is finished), it is possible to get hints to interpret the network behavior that led it to its decisions.

Decompositional approaches have been proposed in the case of pattern recognition \citep{setiono96, remm02}. They include a step of pruning of the network, for the hidden layer to remain in tractable dimension and another step of simplification by mapping the activity of the hidden units onto a finite set of values, corresponding to the features to be extracted. This allows to have a discrete and limited set of combinations to consider, to link input states to the resulting activation of the output, thus yielding rules. This approach has led to sound feature and rule extractions in several application domains. However, when dealing with recurrent networks processing sequences, the extraction task becomes much more difficult.

\subsection{Contexts for implicit learning of sequences}\label{UseCases}

\setlength{\parindent}{1cm}
What is the form of knowledge acquired by learning is a major question in cognitive science and several experimental protocols have been designed to study various paradigms in the case of implicit learning of sequences \citep{perruchet_implicit_2006, pothos_theories_2007}. Two major approaches can be contrasted. 

In the first one, it is believed that contingencies are extracted in a bottom-up way by a kind of associative learning, classical with ANNs and with simple RNNs. In this case, the standard task is the Serial Reaction Time (SRT) task : The subject is asked to reproduce sequences with hidden regularities and improvements of performance in the reproduction of sequences are measured \citep{perruchet_implicit_2006}. 
It is believed that a perceived symbol is associated to the next symbol to produce, and a reduced reaction time is seen as a clue that the transition to be reproduced is valid. In other words, that it has been acquired in the past through the multiple repetitions. The second approach is more top-down and considers that the knowledge is encoded in grammatical rules. The task of Artificial Grammar Learning (AGL) is generally considered here \citep{pothos_theories_2007}. Grammatical sequences built with a finite state language are presented and it is believed that the RNN is able to extract and build an automaton encoding the corresponding language. This process is less dependent on frequences of local transitions and in the test phase, new strings can be classified as grammatical or not grammatical. In all the cases evoked above, a corpus of sequences containing hidden rules is used to train the RNN that will be subsequently analyzed to extract the knowledge implicitly acquired by learning. 

Two contexts for knowledge extraction can appear : When the grammar is known a priori, and when it isn't known. 
In the first context, the extracted rules or contingencies can be directly compared to the original grammar for validation. In this context, it is possible to generate a learning corpus including positive and negative (valid and invalid) examples to learn the task as a categorization problem, or including only positive examples, considering the task as the prediction of the next state. 
In the opposite context, i.e. when the grammar is not known,  assessing the quality of the extracted knowledge is a more difficult process, only possible by experimentation or by the subjective analysis by a human expert. In this context, only positive (valid) sequences are available for learning, suggesting that prediction tasks will be mainly considered here.

In our work, we will mainly consider the AGL approach, i.e when the context is known, so that we will be able to evaluate the results of the proposed methodology. 
But we will discuss also the second context in the results sections by presenting some preliminary work on real data set that involved human evaluation.

\subsection{Existing techniques of knowledge extraction from sequences}\label{existing}



Within the AGL approach, the extracted knowledge implicitly encoded in a recurrent neural network (RNN) can take the form of a finite state automaton (FSA). Automata extraction from RNN can be described as a three-step process \citep{wang2018comparison}, following a decompositional method \citep{remm02}, for post-hoc interpretability. Considering that the network is already trained with a learning corpus, the first step consists, for each item in a test corpus (set of sequences), in recording the values of the hidden units in the shape of vectors. We will refer to them as the hidden units' activation patterns or hidden patterns.
During the second step, all the collected hidden patterns are analyzed to quantize the hidden state space into a finite set of discrete states that can be seen as important features extracted from the network. 
The third step is the automaton construction. 
And finally, the minimization process of the obtained automaton takes place \citep{giles1992learning}. 

\subsubsection{Hidden space quantization}

Among the approaches used to quantize the hidden space, many different approaches were used : 1) considering that each hidden pattern is a state of the desired automaton \citep{weiss2017extracting}, 2) state-space quantization consisting in cuting it into discrete states by choosing a quantization parameter $q$ and keeping only those relevant using a breath-first search (BFS) 
\citep{giles1991second,giles1992learning,omlin1996extraction}
, and 3) clustering techniques like k-means \citep{zeng1993learning}, self-organizing maps \citep{tivno1995learning, blanco2000extracting} and hierarchical clustering \citep{cleeremans1989finite,servan1989learning,elman1990finding}.
One characteristic shared between these previous studies is that they mainly use positive and negative examples generated from the Tomita grammars \citep{tomita1982dynamic}  that induce regular binary languages (sequence of 0 and 1), far from real life choices.
Indeed, only \cite{tivno1995learning} proposed a process for rule extraction from a RNN fed with only positive examples composed with non-binary choices.
Our work is in line with this work, since in real life situations where implicit learning occurs, rules are extracted from learned sequences without any categorization (i.e. there is no positive or negative examples)

\subsubsection{Automata construction}

Many techniques can be used for automata construction once the partitioning is done. They all consists in generating the states and outputs (and the output classification if necessary), as input patterns are fed to the RNN. Indeed, by feeding the network with inputs patterns, it is possible to find the appropriate transitions between the states, and thus the edges of the final automaton \citep{omlin1996extraction}. 
In the case quantification using the parameter $q$ is used, the hidden space is divided into \textit{macrospaces} that represent \textit{"basically
what the rule extraction algorithm “sees” of the underlying RNN"} \citep{jacobsson2005rule}. By feeding the RNN with input patterns, the BFS algorithm  is used to find the appropriate transitions between these macrostates, which correspond to the edges of the final automaton.


Another technique is the sampling-based approach, the principle of which is to replace the search in a quantized state space by the recorded activity of the RNN interacting with its environment and the data \citep{watrous1992induction}. 
\cite{tivno1995learning} proposed a similar approach but only for the extraction of a Mealy machine (where the output depends on the input at time $t$ and the current state) as opposed to a Moore machine (where the output depends only on the current state) \citep{keller2001classifiers}. 
This approach has been extended in more recent studies \citep{wang2017empirical,wang2018comparison}. 
Another approach proposed by \cite{schellhammer1998knowledge}, whose data set was also composed only of positive examples, consists in using the k-means algorithm for clustering the state units activation patterns and a process based on frequency analysis for the automaton construction phase. Only transition whose frequency is above a specific threshold are kept and displayed in the final automaton.
Among all these approaches, we considered that \citep{giles1992learning, omlin1996extraction} are the most consistent ones with the implicit learning phenomena that we wanted to model but applied to only positive examples.  
It is based on the idea that the arrival of a new entry leads to a change of state in the representation of network knowledge. 
It is therefore possible to link network states through these inputs and extract a representation of the implicit rules contained in the data provided as input. In other words, our hypothesis, after learning, is not to extract a complete implicit representation of the network's knowledge, but a representation of the rules as perceived by the network according to the data provided at time t, i.e. only the portion of the set of acquired implicit rules, used to make a prediction.

\subsection{Positioning and main contributions }

The work reported here is concerned with the extraction of structured knowledge acquired by implicit learning when the grammar is unknown and only positive examples are available for learning \citep{tivno1995learning, schellhammer1998knowledge}. 

Our purpose is to model implicit learning and mainly to explicit the implicit rules hidden in the sequences encoded during the learning phase. We thus choose to use a post-hoc interpretability approach. The network is first trained with positive examples i.e. a set of sequences, with the specific task of predicting the symbol at time $t+1$ of the sequence when it receives the symbol at time $t$. Secondly, its hidden activity pattern is then analyzed to explicit its behavior.
To this end, we adapted the algorithms of \citep{giles1992learning, omlin1996extraction}, after the clustering phase, to generate automata using the extracted states and characterized by labelled transitions that reflect the grammatical rules that have been implicitly encoded by the network during its learning phase. 

At the machine learning level, in the models reported above, RNN are generally considered since they are able to process sequences. Basic RNN models have mostly been used for these early attempts of automata extraction. They often demonstrate some weaknesses in the complexity of sequences they can assimilate. More recent RNN models like LSTM have sometimes been considered but, because of the complexity of their inner mechanisms, mostly for theoretical studies and binary grammars. 

In the present paper, we propose a novel methodology for structured knowledge (i.e rules) extraction from LSTM that explicit the behavior of the RNN during the learning of sequences. By expliciting these rules, we propose a methodology in the field of interpretability of RNN that can be applied to different kinds of sequences in contexts where grammar is unknown. Our hypothesis is that the extracted automaton can provide local interpretability about specific predictions, but also provide a substitute of the neural network that can mimic the behavior of the whole network for a specific data set and thus bring hints in the field of global interpretabilty of RNN.

\begin{figure}[htbp]
\centering
\includegraphics[width=\textwidth]{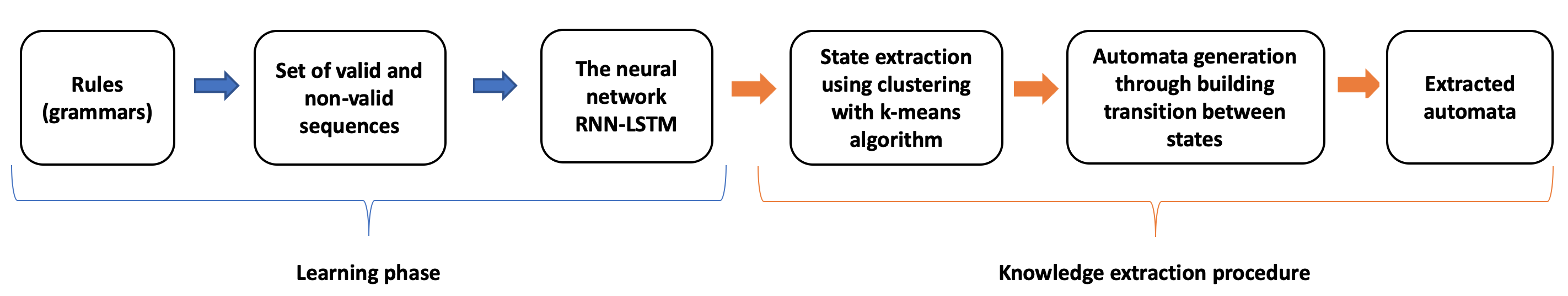}
\caption {\label{fig:LeIA_Process}The global experimental approach for knowledge extraction from RNN LSTM in a task of prediction. Valid and non-valid sequences generated from a grammar are used to train the network. Then valid sequences are used for the extraction of the knowledge implicitly acquired during the learning phase under the shape of an automaton. }
\end{figure}

In the sequel of the paper, we first describe our methodology by describing the LSTM recurrent network used, the knowledge extraction process and the validation process of the extracted knowledge.  
We then introduce
the grammars that have been chosen for generating the corpus of sequences, derived from the Reber grammar, a grammar originally used in cognitive psychology experiments about implicit learning ability in humans \citep{reber1967implicit, cleeremans1991learning}.
Finally, we report corresponding experimental results  before discussing the main lessons about this study and evoking prospective work in the last section. 
Figure \ref{fig:LeIA_Process} depicts the global experimental approach followed in our work.

\section{Methods}


\begin{figure}[htbp]
\centering
\includegraphics[width=\textwidth]{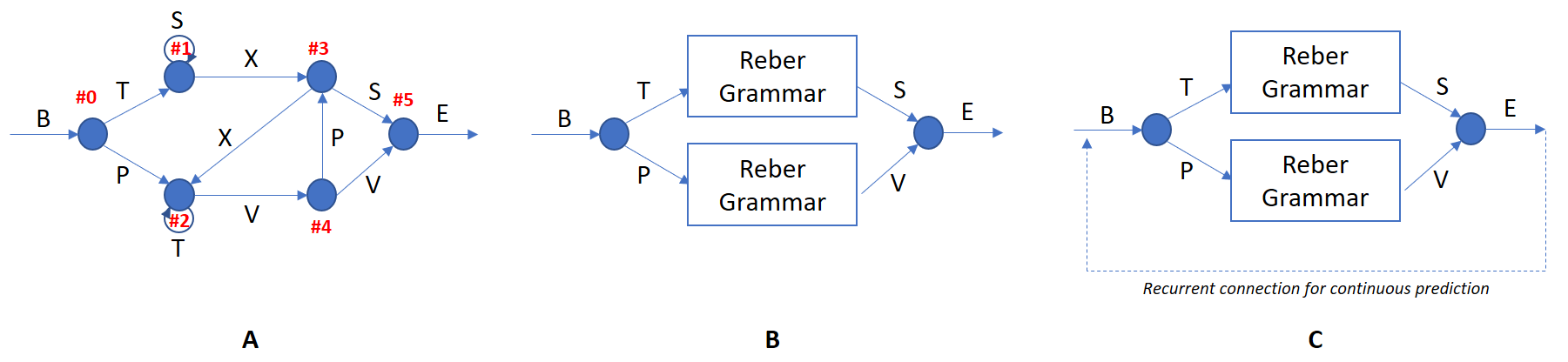}
\caption {\label{fig:RG_ERG_CERG} The three grammars used in the experiments, represented as a Finite State Automaton including nodes representing states and bows emitting symbols. From left to right : A - Reber's Grammar (RG), B - Embedded Reber's Grammar (ERG), C - Embedded Reber's Grammar (CERG). B means "Begin" and E means "End". See text for details.}
\end{figure}


The extraction of implicit rules from recurrent networks was first studied using the Simple Recurrent Network (SRN) \citep{elman1990finding,ServanSchreiber1988encoding, cleeremans1989finite, servan1991graded,cleeremans1991learning}. But it was established that due to the limited size if the context layer, the SRN couldn't handle long term dependencies : the old past is overwritten by the recent past, which makes the development of a stable implicit representation of grammar rules impossible, especially in the case of long and ambiguous sequences.  We thus choose to consider LSTM, among all the existing RNN and architectures, since that model showed great performance for learning sequences from different kinds of dataset \citep{greff2017lstm}. Among the different variants of the LSTM model, we have chosen to implement a neural network using the LSTM units version proposed by \cite{gers1999learning} (with forget gates and no peephole connection).
The network's topology consists in three layers with recurrence limited to the hidden layer composed of four LSTM blocks with two cells each.
The model is trained on the following task : considering a sequence, the network receives as input at time $t$ a symbol and should predict the symbol at $t+1$ in the sequence.
The network is trained and tested on sequences but also flows (sequences put end to end) generated from the variants of the Reber grammar (RG), the embeded reber grammar (ERG) and the continuous and embeded Reber grammar (CERG). All the grammars are presented in figure \ref{fig:RG_ERG_CERG} and described in section \ref{ArtificialGrammars}. Table \ref{Tab_corpusRNN-LSTM} presents the characteristics of the data sets.

\begin{figure}[htbp]
\centering
\includegraphics[width=0.7\textwidth]{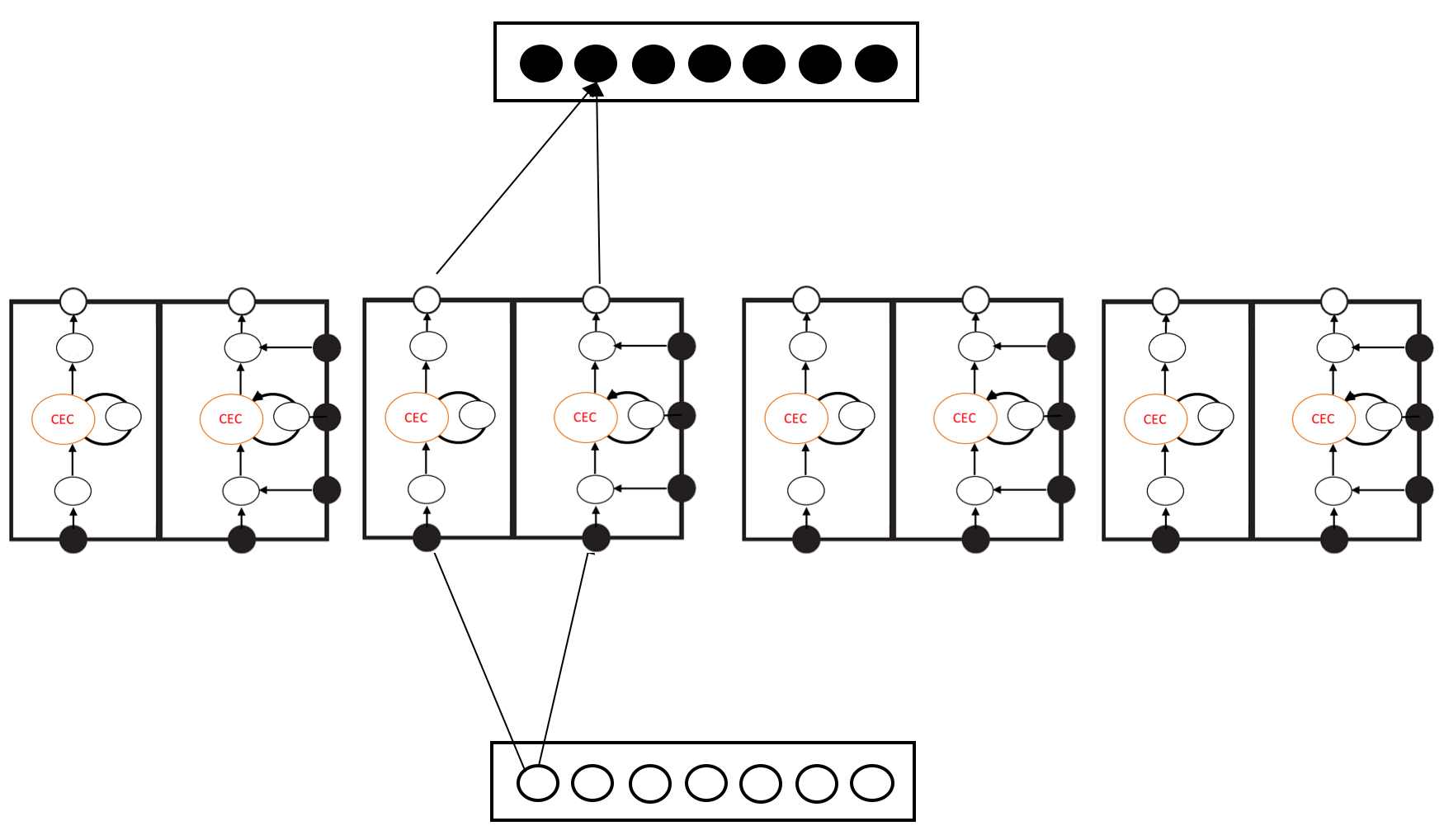}
\caption {\label{fig:RNNLSTM} The RNN-LSTM model with three layers. In addition to classical input and output layers, hidden units are introduced : four LSTM blocks with two cells and a CEC (Constant Error Carousel) in each. In the picture, all white dots outside LSTM blocs are linked to all black dots. There are skipped connections between input and output units. The hidden layer provides a real-valued vector of size 8. Figure adapted from \cite{lapalme2006composition}}
\end{figure}

The first step of our work consisted in reproducing the experimental results of \cite{gers1999learning}. It concludes to the effectiveness of the network to retain long term dependencies and its strength when it deals with very large data corpus (30 000 flows of size 100,000).
The network architecture is represented figure \ref{fig:RNNLSTM}

\subsection{Knowledge extraction procedure}
\label{sec:2_myrulesExtraction}



During the learning phase, it is supposed that important knowledge associated to the hidden rules is encoded in the processing part of the network, at the level of the state space of the hidden layer units. This knowledge should allow to define the rules of transition of the grammar (set of rules) and the contexts in which they are valid. 
Transposing existing approaches for rule extraction from SRNs \citep{ServanSchreiber1988encoding} onto LSTMs, at each time step, we record the activity of the outputs of the LSTM cells, considered here as the activity of the hidden layer. We extract implicitly learned features (corresponding to important contexts for rule definition) by a clustering process onto this hidden space and subsequently build an automaton by collecting sequences of the extracted features when valid sequences are fed to the network. The automaton is minimized in an ultimate step, before its evaluation. 

\subsubsection{Collecting activity patterns of the hidden layer}

To extract the knowledge implicitly learned in the hidden space, we generate a test corpus from the same grammar as provided for training the network (here RG and ERG). To analyze information flow as a sequence is presented, we have created a list of activity patterns, each with a unique identifier consisting of the current symbol and the time step.
In the sequence BTXSE for example, the ids list of the activity patterns for each symbol will be : B0, T1, X2, S3.
We record the activity of the outputs of the LSTM cells that are represented in numerical vectors of dimension 8. 


\subsubsection{Feature extraction algorithm using k-means for clustering}

We use the k-means algorithm for clustering. 
This algorithm consists of partitioning the data collected from the hidden space into k groups by minimizing the distance between samples within each partition.
The number of clusters in which the data must be partitioned is a parameter of the model.
For each value of k, each of the hidden activity patterns ordered in a list that we call "list-patterns" is assigned to 1 among k cluster, thus yielding a list of same size called "list-clusters" indicating the cluster id of each pattern.
Consider two examples of lists: A list-patterns of 5 patterns: [h0, h1, h2, h3, h4] and a list-clusters [0 3 2 2 0] from a clustering using k-means with k = 4.
An appropriate reading of these two lists is: 
\begin{itemize}
    \item the patterns h0 and h4 belong to cluster 0
    \item the pattern h1 belongs to cluster 3
    \item the patterns h2 and h3 belong to cluster 2
\end{itemize}

Matching both lists, it is possible to associate any pattern to a cluster and thus to a state (i.e. a node) in the FSA we wish to extract.

\subsubsection{Average Silhouette coefficient analysis for k-means algorithm}

To analyze the impact of the clustering parameter, the k value, onto the distribution of clusters, we computed the mean silhouette coefficient for all samples for each value of k.
The purpose of the method is to study the distance between resulting clusters on a range of [-1, 1]\citep{rousseeuw1987silhouettes}.
For a given sample, if the silhouette coefficient is close to -1, the sample is assigned to the wrong cluster. If the measure is around 0, the sample is on (or very close to) the decision boundary between two neighboring clusters, and if the measure is close to 1, the sample is assigned to a cluster and far from others. 
The best k value for clustering is reached for the maximum value of the silhouette coefficient. Computing this value allows thus to objectively evaluate the quality of the clustering computed for each value of k.

\subsubsection{Automata generation}\label{automata_generation}


\begin{figure}[htbp]
\centering
\includegraphics[width=\textwidth]{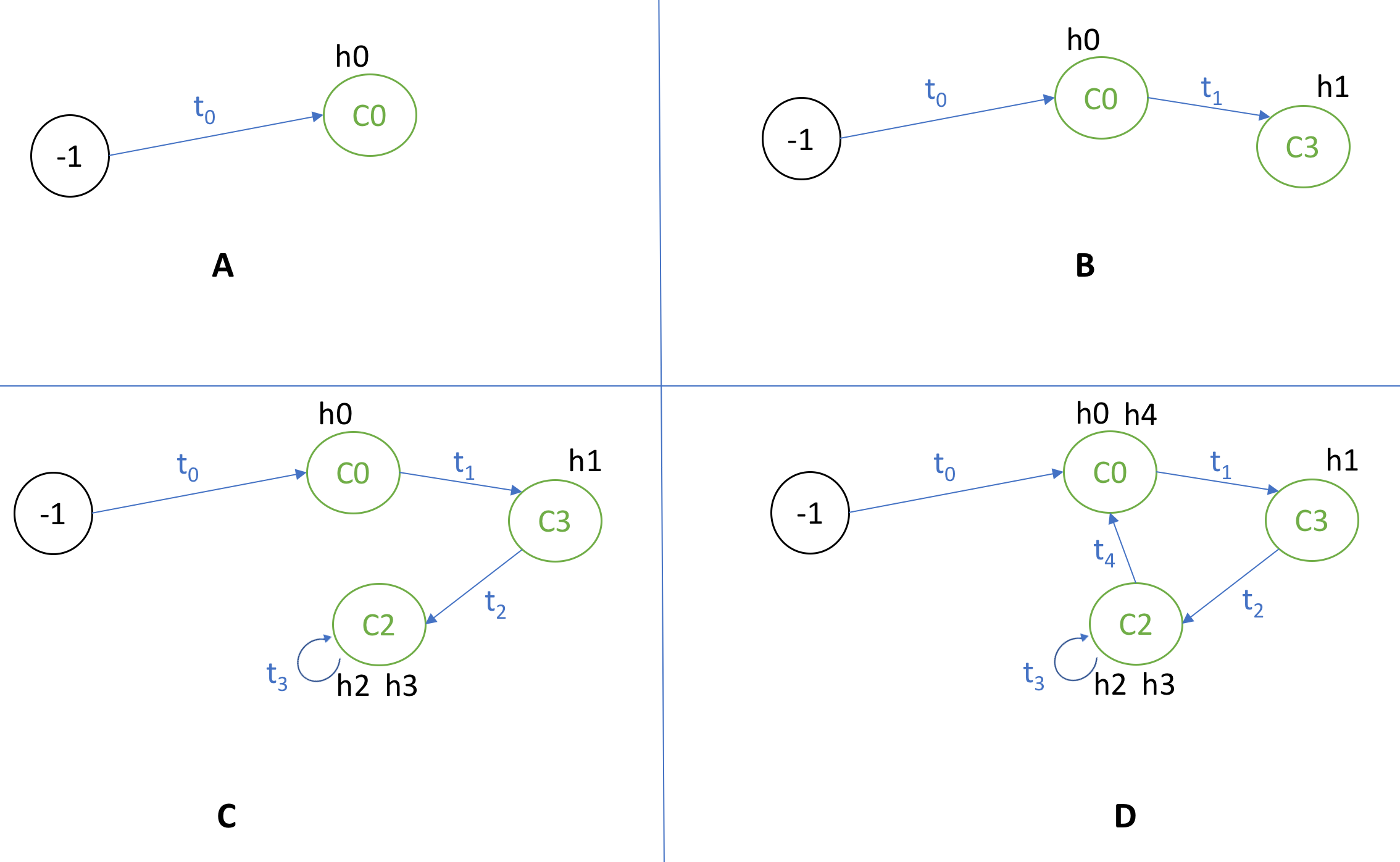}
\caption{ Example of an automaton generation with a list-patterns [h0, h1, h2, h3, h4] and a list-clusters [0 3 2 2 0].
A - Time step $t_0$, creation of a new node (state).
B - Time step $t_1$, creation a second node (state).
C - After time step $t_3$ and $t_4$ : h2 and h3 belong to the same cluster, which generates a loop. 
D - Time step $t_4$ : h4 belongs to a cluster already visited, which creates an edge between the current node (C2) and the existing node (C0)}
\label{fig:RulesExtractionRNN-LSTM}
\end{figure}

We adapted the algorithm described by \cite{omlin1996extraction} for automaton generation.
Figure \ref{fig:RulesExtractionRNN-LSTM} presents an example of automaton generation using the five patterns mentioned above and information given by list-patterns and list-clusters. The generation of the FSA is initiated by adding a node with the identifier -1 as a starting point (figure \ref{fig:RulesExtractionRNN-LSTM}, label A). \\

The rule extraction process needs a simultaneous analysis of both list-patterns and list-clusters : for each pattern, if the associated cluster is a new one (i.e. not represented as a node in the FSA), a new node is then added with the cluster number as its id, together with an oriented edge from the previous node to the new node (figure \ref{fig:RulesExtractionRNN-LSTM}, label A, B, C).
If the current pattern belongs to a cluster already represented in the FSA, then a directed edge is added between the previous node and the corresponding node (figure \ref{fig:RulesExtractionRNN-LSTM}, label D, time $t_4$).
In the case where two consecutive patterns belong to the same cluster, a recursive connection is added to the node representing the cluster (figure \ref{fig:RulesExtractionRNN-LSTM}, label C, time $t_3$). 
This process yields an unlabeled automaton expliciting the arrangement of the clusters (states) of the extracted automaton. To improve the quality of the extracted automaton and give more information on the processing of sequences in the hidden space, we have improved the algorithm proposed in \cite{omlin1996extraction}, also extending the focus made in \cite{schellhammer1998knowledge} on the nodes and their frequency, to edges and the information they carry.

In our approach, each edge is assigned an identifier during the automaton generation phase, corresponding to the symbol that the LSTM processes at the associated time step. 
In the previous example, if sequence BTXSE is the first to be analyzed by the network, the identifier of each symbol will be B0, T1, X2, S3.
If an edge already exists between the two nodes, the new symbol is added to its identifier. 


This original process allows to generate a FSA with long labels on edges, that explicit the temporal organization of the RNN patterns. 
Figure \ref{fig:RG_KMeans_Matrix_DFA_10_labelinput} shows the example of an automaton generated without a label and figure \ref{fig:RG_KMeans_Matrix_DFA2_15} shows the same automaton with labels. The algorithm \ref{Algo_RulesExtractionRNN-LSTM} describes the process of extracting the rules in the form of a FSA with long labels (figure \ref{fig:RG_KMeans_Matrix_DFA2_15}) using the hidden activity patterns of an RNN-LSTM.\\ 

Long labels are interesting because they give information about the place of symbols in the sequence. They can also be synthesized into short ones by just keeping the id of the symbol and removing numerical information. The comparison with the original automaton of the corresponding grammar becomes visually easier (for small automata).

Figures \ref{fig:RG_KMeans_Matrix_DFA3_10_labelinput} and Figures \ref{fig:RG_NOTSimplified_KMeans_Matrix_DFA_10_labelinput} present examples of automata generated with a simplified label. \\

\begin{algorithm}[H]
\begin{algorithmic} 
	
	\REQUIRE 
	 
		\# \textcolor{gray}{Learning and test of the RNN----- }\\
		RNN\_LSTM.learning(learning\_data\_set)\\
		\STATE

		\# \textcolor{gray}{labels\_list : list of symbols presented to the network during tests }\\
		activity\_patterns\_list, labels\_list = RNN\_LSTM.test(test\_data\_set)\\
	
	\STATE

	\STATE \textbf{Function rules\_extraction (activity\_patterns\_list, labels\_list,  k)} : \\

                \STATE {
	        		\# \textcolor{gray}{Clustering  ---------------------------------------------------- }\\
	        		clusters\_list = k\_means(k, activity\_patterns\_list) \\
			\STATE
			\STATE
			\# \textcolor{gray}{Generation of automaton A---------------------------------------------------- }\\
			A = \COMMENT {} \# \textcolor{gray}{Dictionnary}\\
			current\_node= -1\\
            
			A['nodes'].add(current\_node)\\
			A['edges'] = [ ]  \# \textcolor{gray}{list of dictionnaries}\\
			
			\FORALL{pattern $h$ of index $i$ from activity\_patterns\_list}
				\STATE associated\_cluster =  clusters\_list[$i$]
					\IF{associated\_cluster $\not \in $ A['nodes']} 
						\STATE{A['nodes'].add(associated\_cluster)} 
					\ENDIF \\
					
					\STATE
					edge= \COMMENT {} \# \textcolor{gray}{Dictionnary}\\
					edge['id'] =  (current\_node, associated\_cluster)\\
					\IF { edge $\not \in $ GA['edges'] } 
						\STATE{
							new\_edge = edge\\
							new\_edge['weight'] = 1\\
							new\_edge['label'] = labels\_list[$i$]\\
							A['arcs'].add(nouvel\_edge)
							} 
					\ELSE 
						\STATE{
							edge['weight'] = edge['weight'] +1\\
							edge['label'] = edge['label']+ labels\_list[$i$]\\
							A['edges'].update(edge)\\
						}
					\ENDIF\\

					\# \textcolor{gray}{Update of the current node}\\
					current\_node= associated\_cluster
				
			\ENDFOR
			
			\STATE
			
			\RETURN A
			
			\STATE
			}
	\STATE \textbf{End Function}

\end{algorithmic}
\caption{Algorithm for extracting rules in the form of a NFA with long labels, using the activity patterns of an RNN-LSTM \label{Algo_RulesExtractionRNN-LSTM}}
\end{algorithm}

\subsubsection{Automaton minimization}\label{section_automata_minimization}

The last step for automaton generation is the minimization process.
Minimization algorithms exist for deterministic finite automaton (DFA). They consist in transforming a DFA into a minimal version of that DFA with the minimum number of states.
The process starts by determining the type of the FSA : Non-deterministic finite automaton (NFA) or DFA. A DFA has transitions that are uniquely determined by the input symbol from each state. On the contrary, a NFA is an automaton where several possibilities of transition can exist simultaneously for a state and a given input symbol. In this case, a conversion into a complete DFA is made.

In our work we use the table-filling method, also known as the Myhill-Nerode theorem \citep{nerode1958linear}, for the minimization process. 
This method consists in first building a transition table containing as many columns as lines, and where each of them represents the states of the automaton.
The second step is an iterative process that consists of filling the table according to the transitions of the automaton.
The remaining unmarked pairs are grouped as a single state.
Algorithm \ref{Algo_Myhill-Nerode_DFA_Minimization} describes this algorithm.

\begin{algorithm}[H]
\begin{algorithmic} 
	\STATE \textbf{Function rules\_extraction (final\_states, non\_final\_states)} : \\
            \STATE 
            	{Draw a table for all pairs of states (P, Q)\\
            
	        	Mark all pairs where only one state belongs to the final\_states\\
	        	Repeat until no more marking can be done :\\
	        	    \STATE
	        	       \bindent
	        	        \IF{there is any unmarked pairs (P,Q)}
	        	            \bindent
	        	            \STATE
	        	            \# \textcolor{gray}{$\delta$ is the transition function that "transform P into P' according an input X }\\
    	        	        \STATE
    	        	        \IF{[$\delta(P,X),\delta(Q,X)]$ is  marked}
    	        	           \STATE Mark(P,Q)\\
    					    \ENDIF 
    					    \eindent
					    \ENDIF
					    \eindent
	        }
	\STATE \textbf{End Function}
\end{algorithmic}
\caption{The Myhill-Nerode algorithm for DFA minimization \label{Algo_Myhill-Nerode_DFA_Minimization}}
\end{algorithm}


\subsection{Validation procedure and analysis}

\begin{figure}[htbp]
\centering
\includegraphics[width=0.7\textwidth]{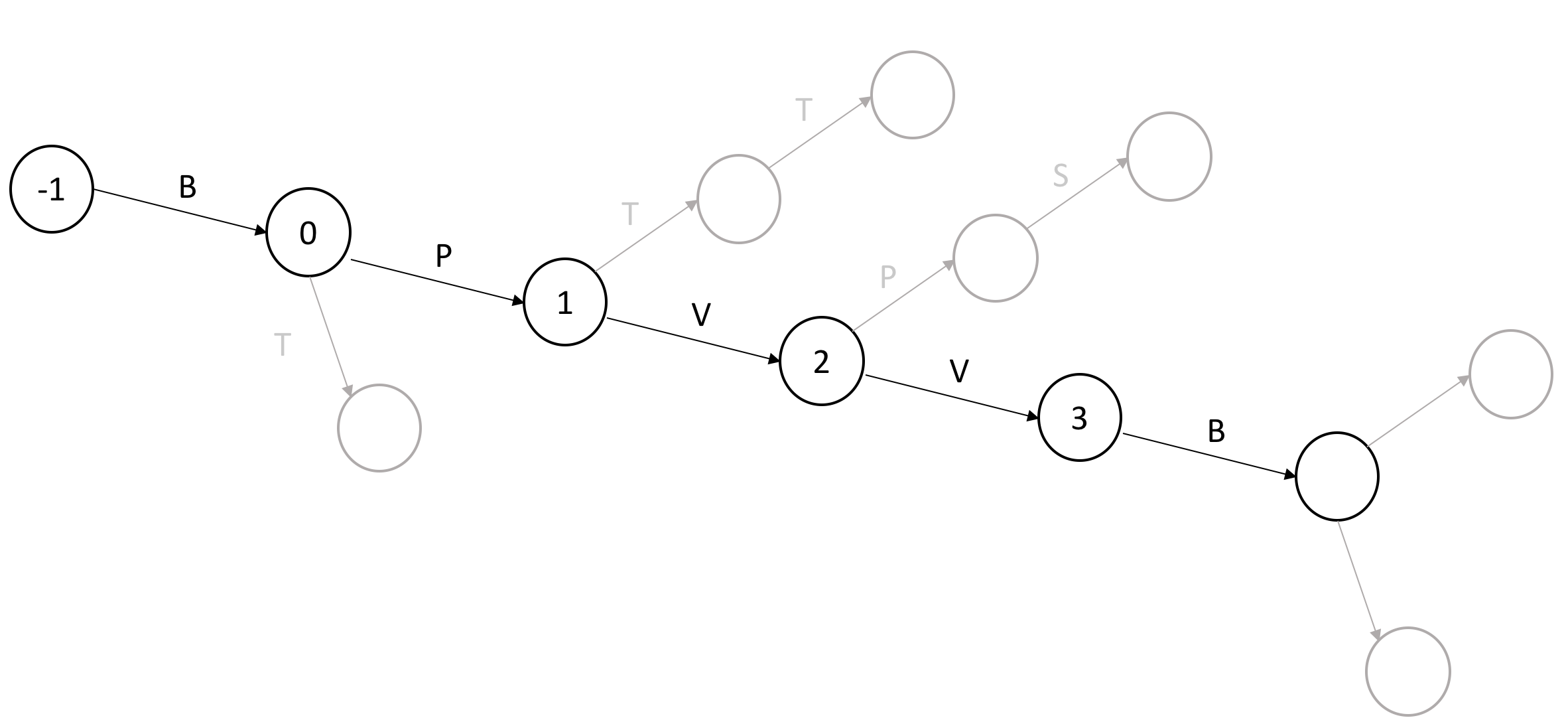}
\caption{\label{fig:ValidationSequencesProcess} Schematic representation of the testing process of the original sequence BPVVE from the Reber Grammar on the extracted and minimized DFA. In black the selected path on the minimized DFA corresponding to the sequence, in gray the ignored ones.}
\end{figure}

The last step of the proposed methodology consists in analyzing the minimized DFA by feeding it with grammatical sequences in order to confirm if it recognizes the same language as the original grammar.
For this purpose, we apply the following process (figure \ref{fig:ValidationSequencesProcess}) : 
For each sequence, the starting node is the one containing -1 in its label.
We apply on it the input (corresponding to the first symbol of the sequence) to retrieve a new state. 
We establish then the neighbors list (i.e states) of this new state and their associated transitions.
If among these transitions, there is one corresponding to the next symbol of the sequence, the new state becomes the current state. The process is then repeated again, until the next symbol of the sequence is the last symbol of the sequence (i.e symbol E).
In this case, we check if among the successors there is the beginning symbol (i.e symbol B, figure \ref{fig:RG_NOTSimplified_KMeans_Matrix_DFA_10_labelinput}). If this condition is true, the minimized DFAs not only recognizes the full sequence, but also respects the long term dependencies of the original grammar.
If among the transitions of the neighbors, none of them corresponds to the desired next symbol, the sequence is rejected.
For each k value for the clustering, we measured the average percentage of accepted sequences for 10 simulations on different grammatical sequences. 

\section{Results}
\label{Exp_Results}

All the simulation were run on a Macbook Pro using the Python scientific stack, namely Numpy \citep{Walt:2011}, Scipy \citep{Jones:2001}, Matplotlib \citep{Hunter:2007} and scikit-learn \citep{scikit-learn}. Sources are
freely available on Github at \textbf{ADDRESS NEEDED}.





\subsection{Artificial data set}
\label{ArtificialGrammars}

Three grammars were used during this work: the grammar of Reber initially proposed in 1967 \citep{reber1967implicit} in cognitive psychology experiments, as well as two of its variants. Each one was employed to test a particular aspect of the RNN, namely performances during prediction, resistance to the size of the sequences and good retention of information when long term dependencies between elements are to be considered.
Based on these grammars, two kinds of sequences will be considered : 
Sequences possibly generated by the automaton (or following the transition of the grammar) are called valid sequences.
Other sequences are called non-valid sequences (i.e, not respecting the grammar).
Figure \ref{fig:RG_ERG_CERG} presents the grammars, table \ref{Tab_corpusRNN-LSTM} the characteristics of the data set and table \ref{Tab_SeqRG} provides examples of both types of sequences.

As it is often the case in AGL, even if the grammar are known, we will consider here tasks of implicit learning corresponding to learn an unknown grammar. 
Only valid (i.e. grammatical) sequences are considered in the learning phase, with no other knowledge about the grammar: These valid sequences might have been collected directly from the environment as it is the case in natural language processing. In a test phase, valid sequences will be used to evaluate more precisely the quality of the extracted FSA but will not participate to its elaboration.

\begin {table}[htbp]
\centering
  \begin {tabular} {| l | c | c |}
  \hline & RG & ERG \\
  \hline
  \multicolumn {3} {| l |} {\textbf{Learning Corpus}} \\
  \hline Size & 200,000 & 200,000 \\
  Number of samples & 1,397,109 & 2,198,671 \\
  Average length & 7.98 & 11.99 \\
  Standard deviation & 3.35 & 3.37 \\

  \hline
  \multicolumn {3} {| l |} {\textbf{Corpus of valid test sequences}} \\
  \hline Size & 20,000 & 20,000 \\
  Average number of samples & 140 193 & 219 411 \\
  Average length & 8.02 & 11.97 \\
  Standard deviation & 3.47 & 3.35 \\
  \hline
  \multicolumn {3} {| l |} {\textbf{Corpus of non-valid test Sequences}} \\
  \hline Size & 130,000 & 130,000 \\
  Average number of samples & 140,538 & 219,533 \\
  Average length & 8.00 & 7.01 \\
  Standard deviation & 6.51 & 5.50 \\
  \hline
  \end {tabular}
\caption {Characteristics of the data sets. Figures concerning the test set are averages calculated on 10 corpuses.} \label{Tab_corpusRNN-LSTM}
\end {table}

\begin {table}[htbp]
\centering
\begin {tabular} {| l l |}
\hline Valid sequences & Non-valid sequences  \\
\hline
BTXSE & BE\\
BPVVE & BVPXE \\
BTSXXVPSE & BTPPPPPE\\
BPTTTTVPSE & BPSE \\
BTSSSSSSSSSSSXSE & BSPPTTTTTTTTE\\
\hline
\end {tabular}
\caption{Examples of valid sequences generated from Reber's grammar and non-valid sequences generated using Reber's symbols} \label{Tab_SeqRG}
\end{table}



At the experimental level, two different contexts were studied : RG and ERG context.
In each context, a learning phase of 200 000 sequences is made, followed by 10 simulations of test phase. For each simulation, we recorded the hidden patterns for 20 000 sequences.

\subsubsection{FSA Extraction : visual results for interpretability}

\begin{figure}[htbp]
\centering
\begin{subfigure}{.3\textwidth}
  \centering
  \includegraphics[width=0.50\textwidth]{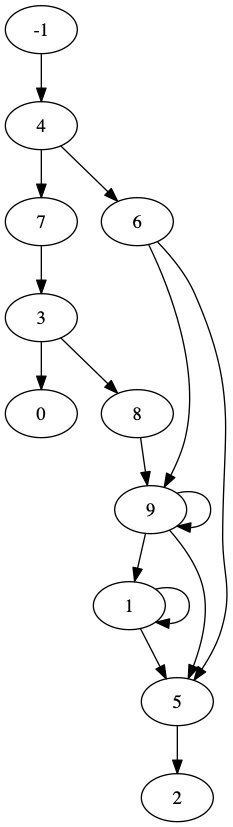}
  \caption{}
  \label{fig:RG_KMeans_Matrix_DFA_10_labelinput}
\end{subfigure}%
\begin{subfigure}{.3\textwidth}
  \centering
  \includegraphics[width=0.60\textwidth]{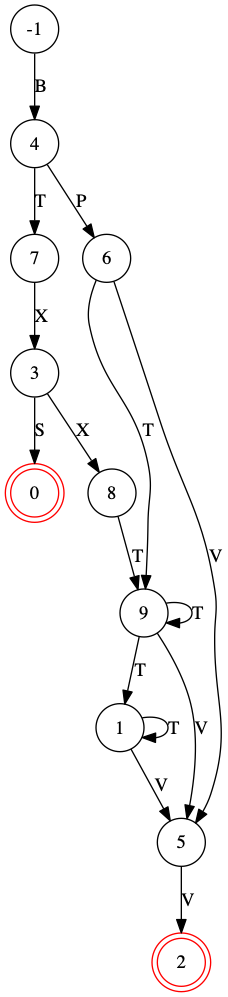}
  \caption{}
  \label{fig:RG_KMeans_Matrix_DFA3_10_labelinput}
\end{subfigure}
\begin{subfigure}{.3\textwidth}
  \centering
  \includegraphics[width=0.60\textwidth]{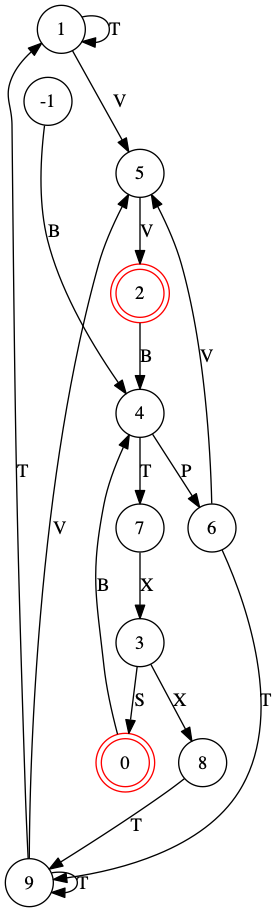}
  \caption{}
  \label{fig:RG_NOTSimplified_KMeans_Matrix_DFA_10_labelinput}
\end{subfigure}%
\caption{Extraction in RG context : An unlabeled FSA (a) and a final FSA (b), obtained with a k-means algorithm for clustering, where k=10.
Figure (c) is the original extracted FSA where sequences do not start from the starting node -1. Final nodes, that indicate the end of sequences (i.e that the following symbol is E), are noted with red double circles.}
\label{fig:RG_KMeans_Matrix_DFA_DFA3_10_labelinput}
\end{figure}

All figures in this section present FSA before minimization extracted from a small number of hidden patterns. In figures \ref{fig:RG_KMeans_Matrix_DFA_10_labelinput} and \ref{fig:RG_KMeans_Matrix_DFA3_10_labelinput}, we represent a simplified version of the graph, without the loops labelled with the symbol B from the final node to the starting node.

\begin{figure}[htbp]
\centering
\includegraphics[width=0.30\textwidth]{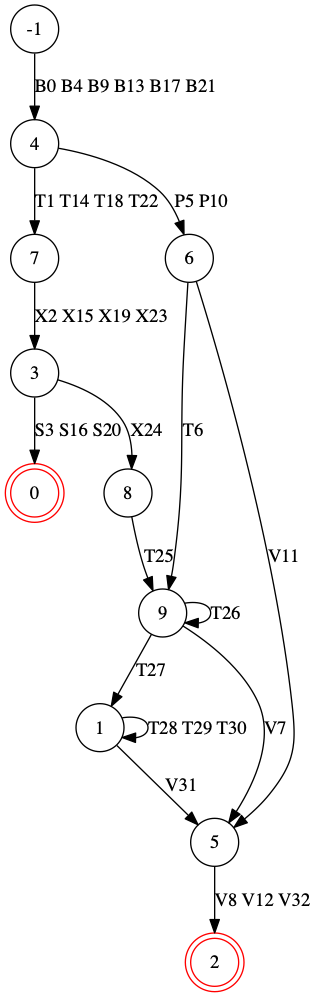}
\caption{\label{fig:RG_KMeans_Matrix_DFA2_15} Extraction in RG context : A long-label FSA obtained with a k-means algorithm for clustering, where k=10 on the same data used in figure \ref{fig:RG_KMeans_Matrix_DFA_DFA3_10_labelinput}.}
\end{figure}

In the RG context, by following the methodology described previously in section \ref{automata_generation}, we obtain, for k=10 clusters, the FSA without label represented in figure \ref{fig:RG_KMeans_Matrix_DFA_10_labelinput}, the FSA with long labels represented in figure \ref{fig:RG_KMeans_Matrix_DFA2_15} and the final automaton graph in figure \ref{fig:RG_KMeans_Matrix_DFA3_10_labelinput}.
These results were extracted from the analysis of the first 30 time steps (i.e the first 30 hidden patterns recorded) during the test phase using a data set composed of various occurrences of the following sequences : BPVVE, BTXSE, BPTVVE, BTXXTTTTVVE.

\begin{figure}[htbp]
\centering
\begin{subfigure}{.5\textwidth}
  \centering
  \includegraphics[width=0.50\textwidth]{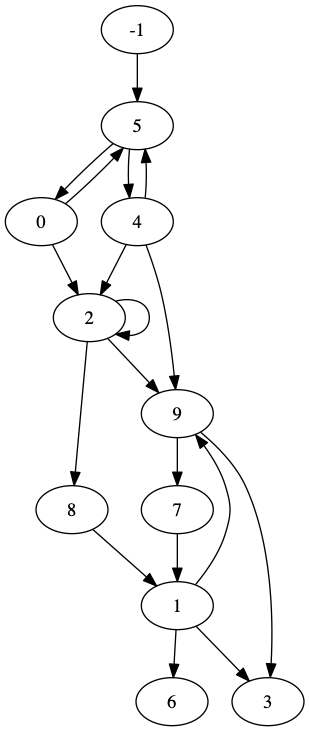}
  \caption{ }
  \label{fig:sub1_ERG}
\end{subfigure}%
\begin{subfigure}{.5\textwidth}
  \centering
    \includegraphics[width=0.50\textwidth]{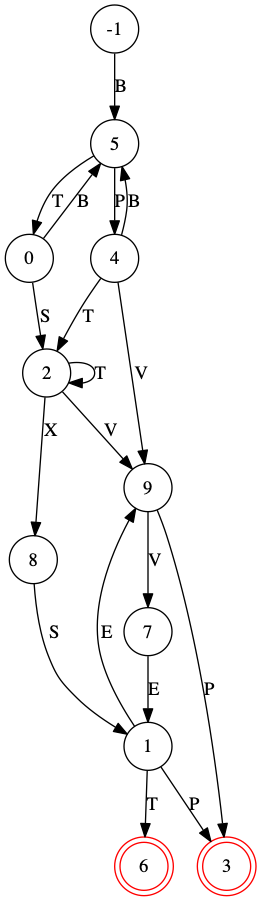}
  \caption{ }
  \label{fig:sub2_ERG}
\end{subfigure}
\caption{Extraction in ERG context : An unlabeled FSA, on the left (a), and a final FSA, on the right (b), obtained with a k-means algorithm for clustering, where k=10. Final nodes are noted with red double circle.}
\label{fig:ERG_KMeans_Matrix_DFA_and_DFA3_10}
\end{figure}

\begin{figure}[htbp]
\centering
\includegraphics[width=0.40\textwidth]{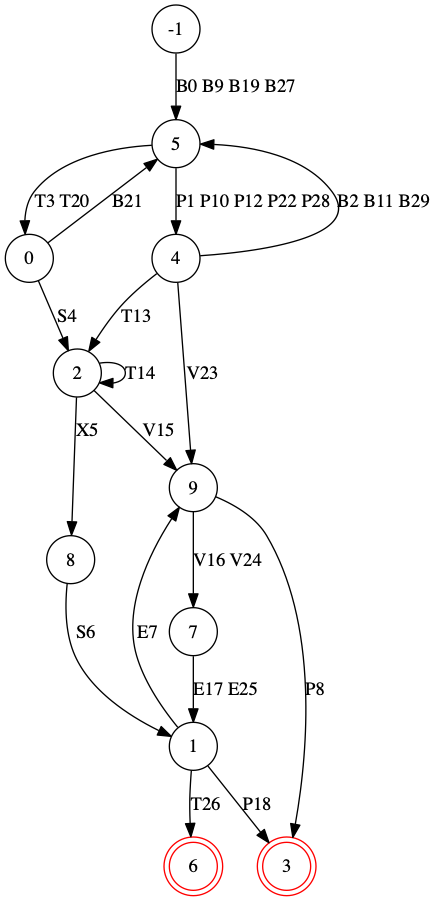}
\caption{\label{fig:ERG_KMeans_Matrix_DFA2_19}
Extraction in ERG context : A long-label FSA obtained with a k-means algorithm for clustering, where k=10 on the same data used for figure \ref{fig:ERG_KMeans_Matrix_DFA_and_DFA3_10}
}
\end{figure}

In the ERG context, an extraction process was realized on 80 hidden patterns related to several occurrences of the 4 sequences: BPBPTVVEPE, BTBPVVETE, BPBTSXSEPE, BPBPTTVVEPE.
We thus obtain, for k=10 clusters, one FSA with 3 different notation systems: the FSA without label, represented in figure \ref{fig:sub1_ERG}, the FSA with long labels represented in figure \ref{fig:ERG_KMeans_Matrix_DFA2_19} and the final FSA in figure \ref{fig:sub2_ERG}.

\begin{figure}[htbp]
\centering
\includegraphics[width=0.5\textwidth]{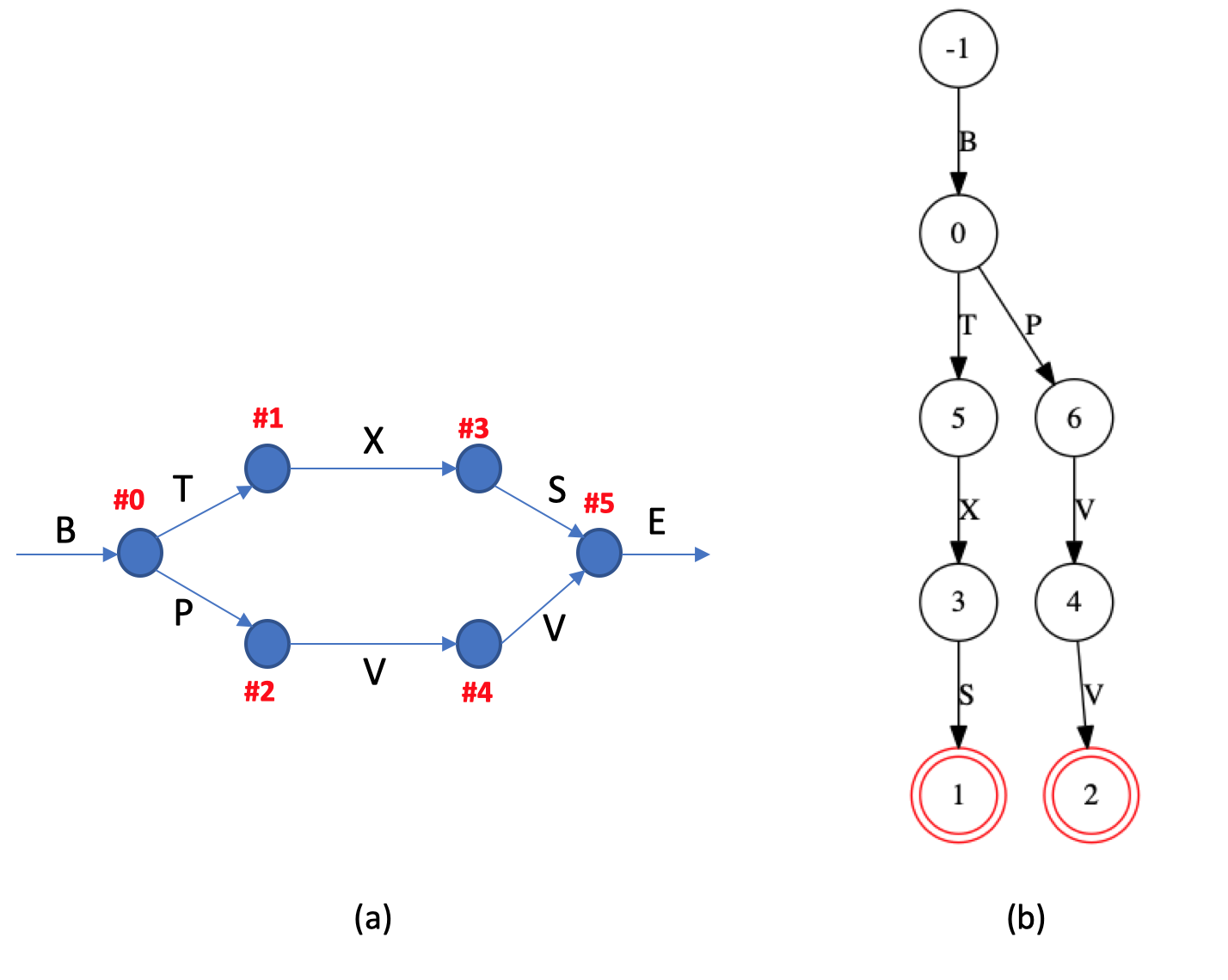}
\caption{\label{fig:GramRG_graphRG} Comparison of a portion of Reber's grammar (a) and an extracted FSA for k=9 (b) for the 15 first time steps related to occurrences  of  2  sequences: BPVVE  and BTXSE. }
\end{figure}

In the case of testing the model on a small volume of data, the extracted FSA will not represent all the implicit and encoded representation of all the learnt data, but just the part of that representation that corresponds to those inputs.
This is why in figures  \ref{fig:RG_KMeans_Matrix_DFA_DFA3_10_labelinput}, \ref{fig:RG_KMeans_Matrix_DFA2_15}, \ref{fig:ERG_KMeans_Matrix_DFA_and_DFA3_10}, \ref{fig:ERG_KMeans_Matrix_DFA2_19} and \ref{fig:GramRG_graphRG} some of the loops and the transitions that are originally present on the RG and ERG may be absent.

\begin{figure}[htbp]
\centering
\includegraphics[width=0.6\textwidth]{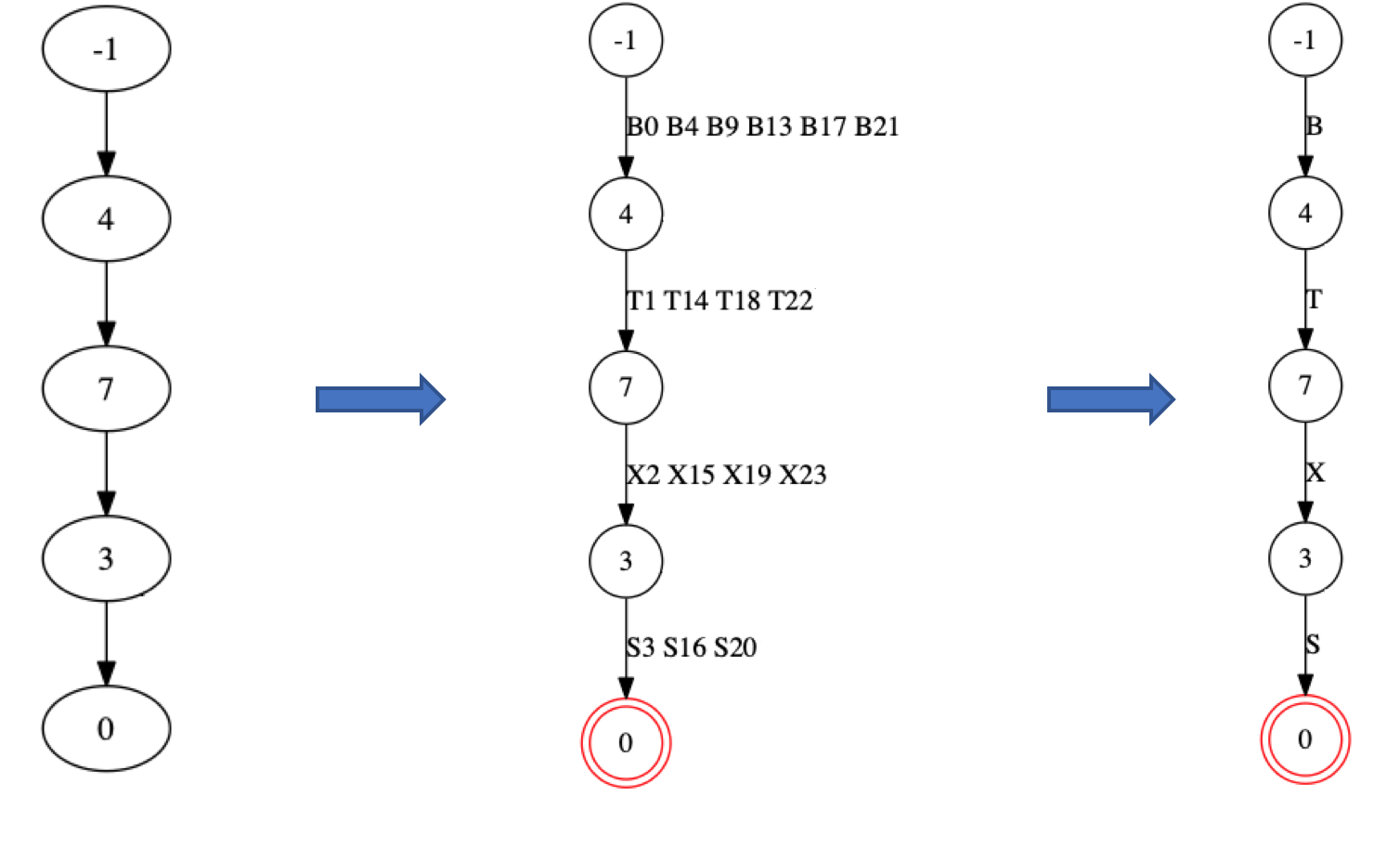}
\caption{\label{fig:RulesExtracted3graphes}
Rules extracted from an RNN-LSTM having learned RG sequences. Examples from the FSA without labels (figure \ref{fig:RG_KMeans_Matrix_DFA_10_labelinput}), the FSA with long labels (figure \ref{fig:RG_KMeans_Matrix_DFA3_10_labelinput}) and the FSA with a simple label (figure \ref{fig:RG_KMeans_Matrix_DFA_DFA3_10_labelinput}) for k=10.} 
\end{figure}

\begin{figure}[htbp]
\centering
\begin{subfigure}{.6\textwidth}
  \centering
  \includegraphics[width=\textwidth]{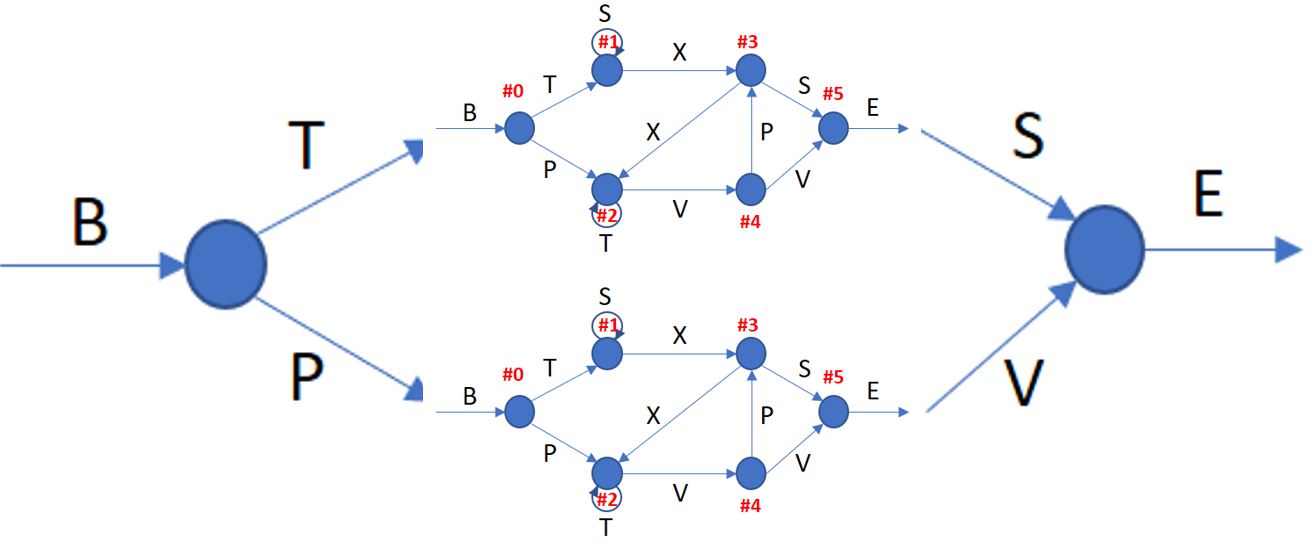}
  \caption{ }
  \label{fig:sub1_ERG}
\end{subfigure}%
\begin{subfigure}{.4\textwidth}
  \centering
    \includegraphics[width=0.40\textwidth]{ERG_KMeans_Matrix_DFA3_10_labelinput.png}
  \caption{ }
  \label{fig:sub2_ERG}
\end{subfigure}
\caption{Comparaison of ERG (a) and an extracted FSA for k=10 (b)}
\label{fig:GramERG_graphERG}
\end{figure}

The main result to underline in this section is the visual interpretability that the labels on transition provide.
In figures \ref{fig:RulesExtracted3graphes}, \ref{fig:GramRG_graphRG} and \ref{fig:GramERG_graphERG}, we present a comparison between the original grammars and the extracted FSA which represent the implicit representation as encoded by the network. On top of that, figures \ref{fig:RG_KMeans_Matrix_DFA2_15}, \ref{fig:ERG_KMeans_Matrix_DFA_and_DFA3_10} explicit both temporal behavior of the model. In other words, these knowledge representations allow to explicit the reasons of the behavior of the network at a specific time step. It is thus valuable for local interpretability in RNN.

\begin{figure}[htbp]
\centering
\includegraphics[width=0.15\textwidth]{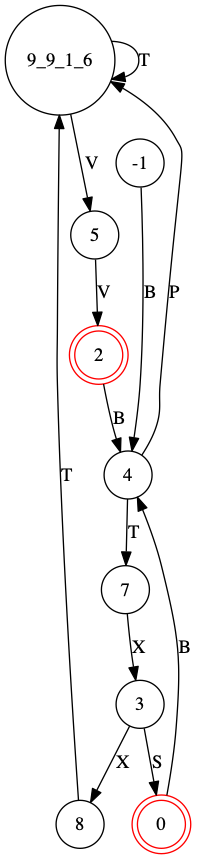}
\caption{\label{fig:Minimized_DFA_Table_filling_method_nbCluster10_labelinput_light}Light version of minimized DFA for k=10 of the final FSA on figure \ref{fig:RG_KMeans_Matrix_DFA3_10_labelinput}. For more clarity, the edges between the different nodes and the "trash node" -2 aren't represented. The node 9\_9\_1\_6 results from the merge of the nodes 9\_1, 9 and 6, due to the minimization algorithm.}
\end{figure}

\begin{figure}[htbp]
\centering
\begin{subfigure}{.5\textwidth}
  \centering
  \includegraphics[width=\textwidth]{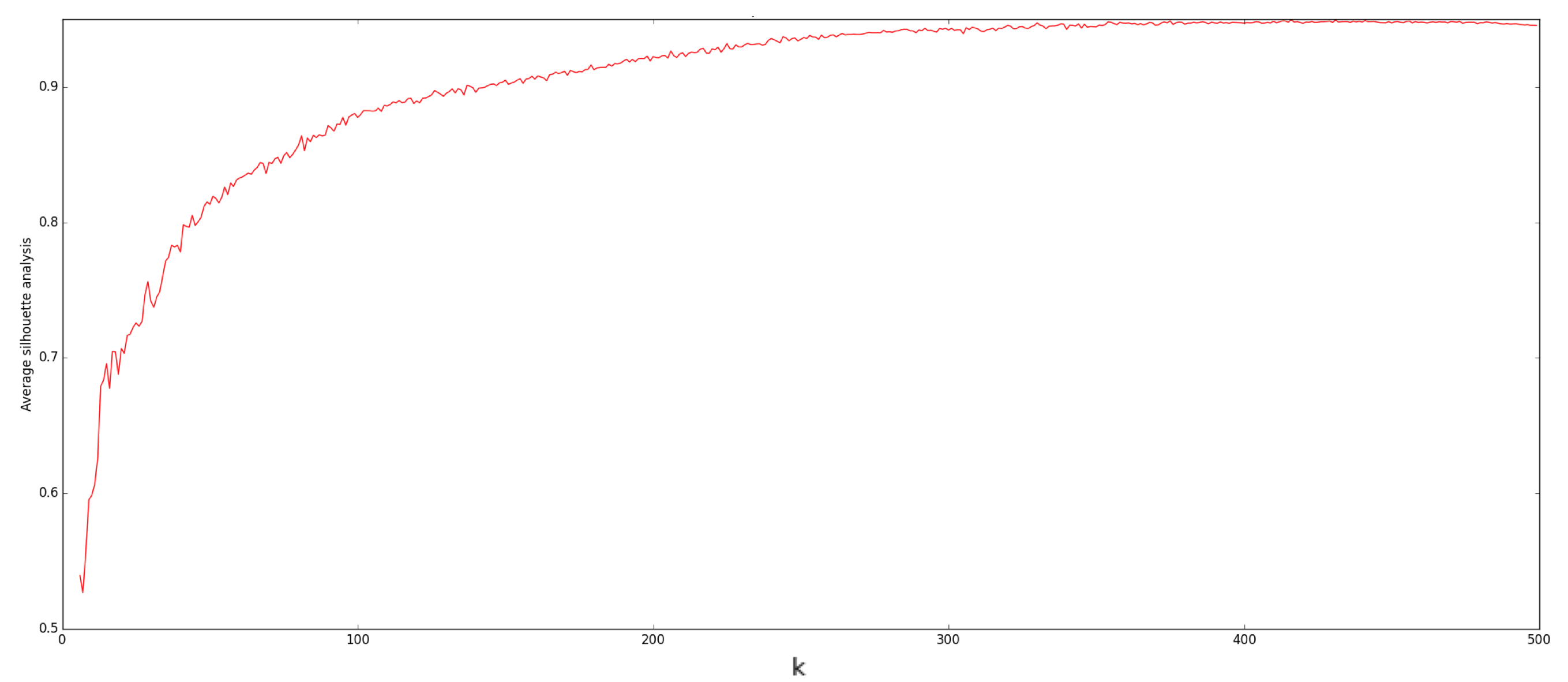}
  \caption{}
  \label{fig:RG-k-evolution}
\end{subfigure}%
\begin{subfigure}{.5\textwidth}
  \centering
  \includegraphics[width=\textwidth]{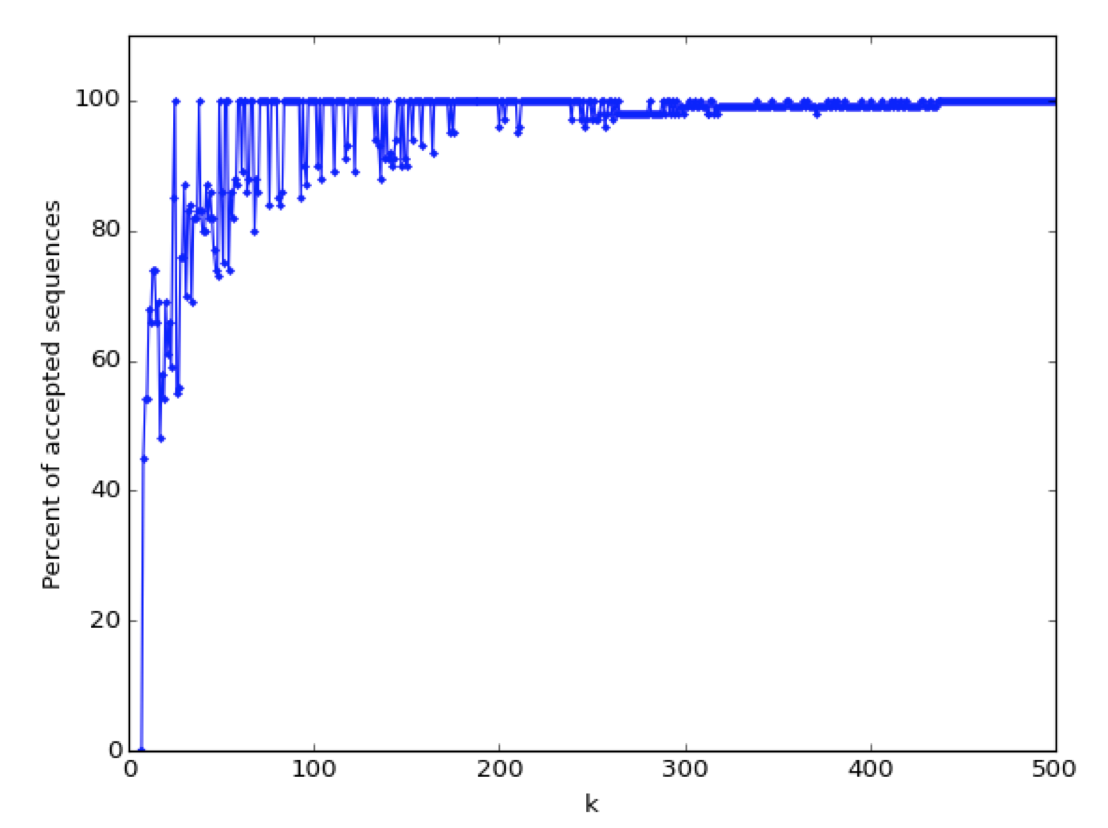}
  \caption{}
  \label{fig:RG-accepted-minimized-DFA}
\end{subfigure}
\caption{Analysis of extraction process of 5000 patterns in RG context for k $in$ [6,500] :  Evolution of the average silhouette coefficient (a) and percentage of recognized sequences from RG (b).}
\label{fig:RG-analysis-minimization}
\end{figure}

\begin{figure}[htbp]
\centering
\begin{subfigure}{.5\textwidth}
  \centering
  \includegraphics[width=\textwidth]{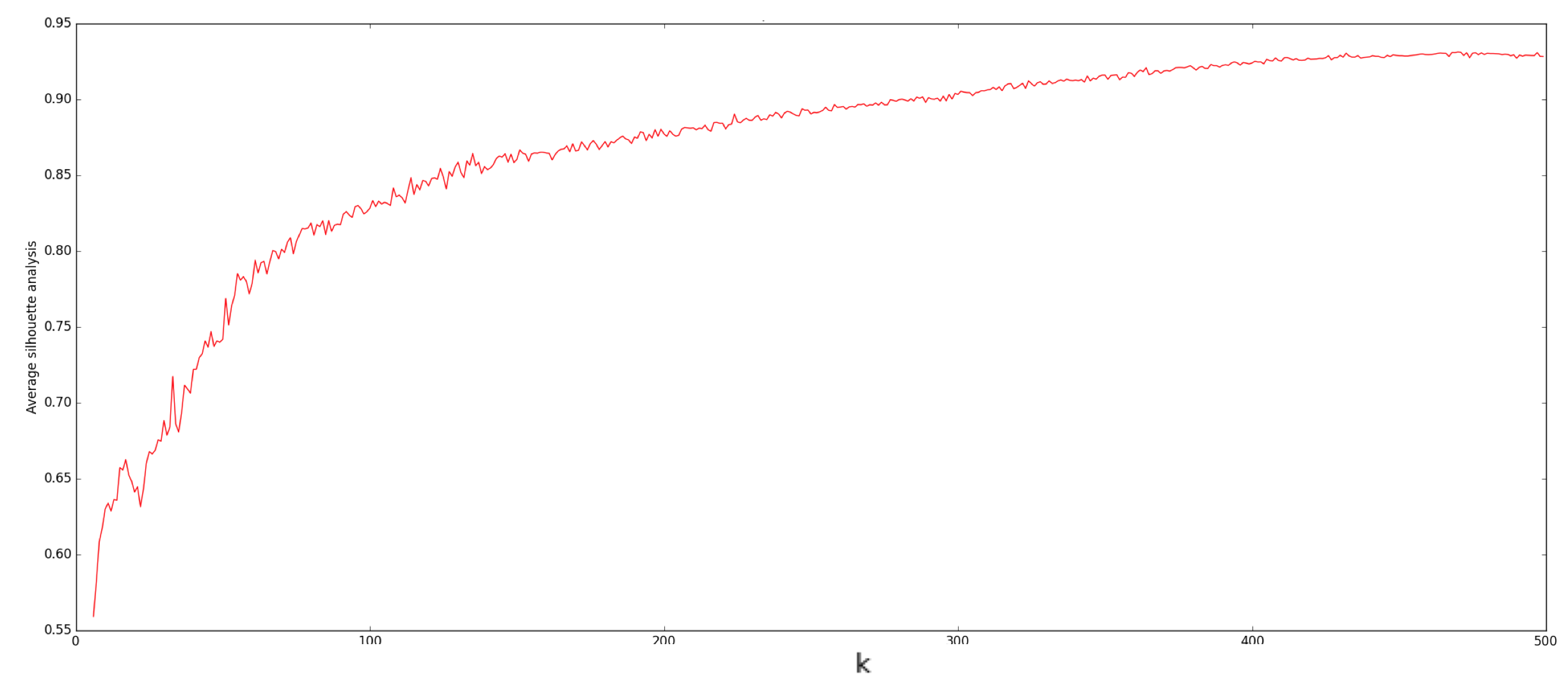}
  \caption{}
  \label{fig:ERG-k-evolution}
\end{subfigure}%
\begin{subfigure}{.5\textwidth}
  \centering
  \includegraphics[width=\textwidth]{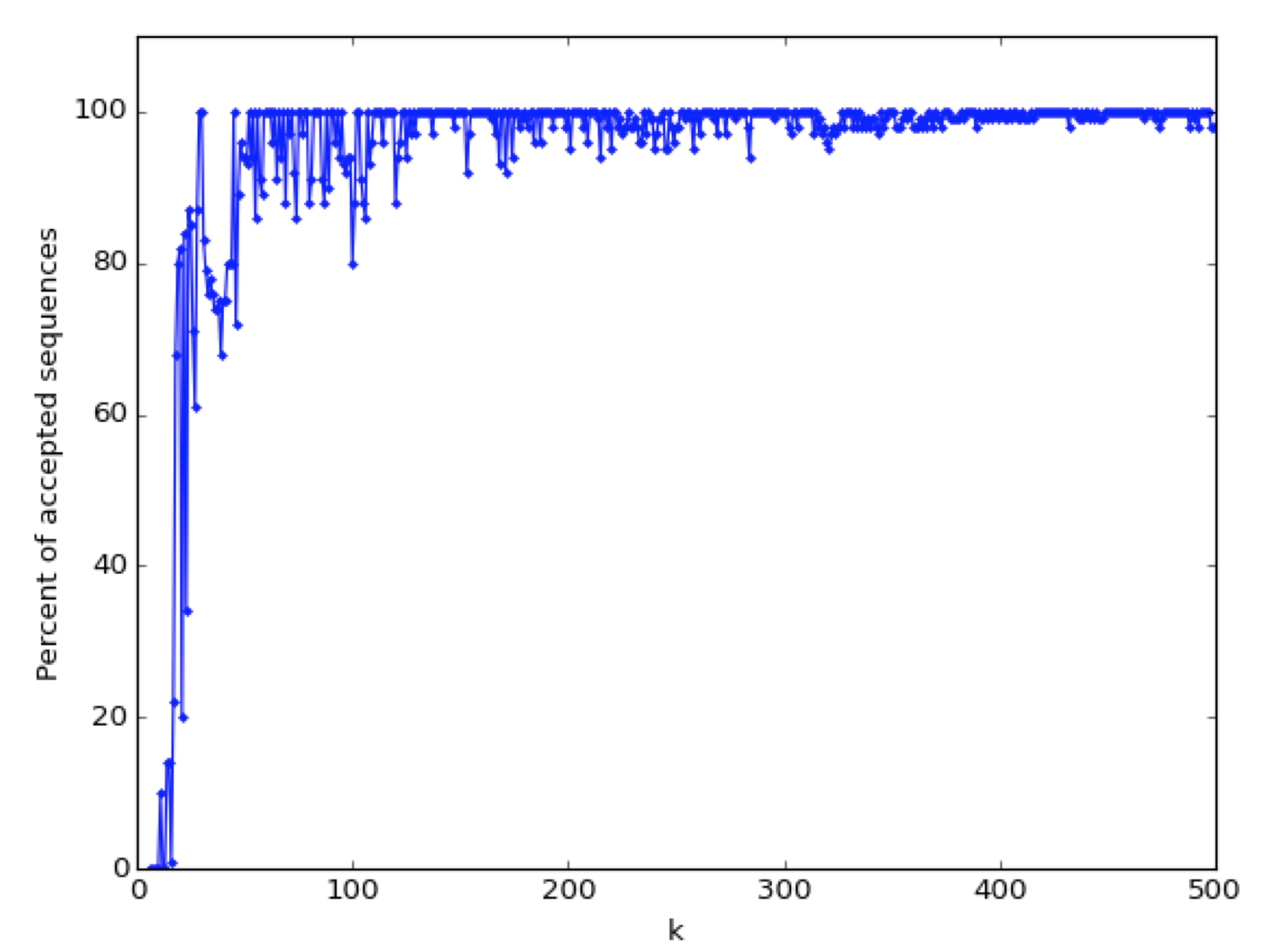}
  \caption{}
  \label{fig:ERG-accepted-minimized-DFA}
\end{subfigure}
\caption{Analysis of extraction process of 5000 patterns in ERG context for k $in$ [6,500] : Evolution of the average silhouette coefficient  (a) and percentage of recognized sequences from ERG (b).}
\label{fig:ERG-analysis-minimization}
\end{figure}

\subsubsection{FSA Extraction : evaluation of the extracted FSA using valid sequences}

To evaluate the quality of the results, we analyze: 1) the evolution of the average silhouette coefficient for each k value (only the results of analysis of the first 5000  patterns are shown, unless stated otherwise) and 2) the percentage of valid sequences recognized by each minimized DFA for each k value. 

Figures \ref{fig:RG-analysis-minimization} and \ref{fig:ERG-analysis-minimization} represent, for each value of k, the evolution of the average silhouette coefficient and the percentage of recognized valid sequences in respectively RG and ERG contexts.
Even if the average silhouette coefficient values decrease only for k in [400, 500], the minimized DFA in both contexts recognized more than 70\% of the valid sequences from k $>$ 50.
If we compare the evolution of the percentage of accepted sequences to the silhouette coefficient analysis of k in both grammars, it appears that the higher the silhouette coefficient, the more it is possible to get an extracted minimized DFAs that can recognize the original sequences, but also approach the original rules (and thus the original grammar) hidden in those sequences.\\

We repeated the extraction process on 5000 different patterns 10 times, and we obtained the following results (average values) :

For RG context : 
    \begin{itemize}
        \item for k $>$ 50, average value of silhouette coefficient is $>$ 0.80 and the percentage of accepted sequences is between [70; 100] 
        \item for k $>$ 100, average value of silhouette coefficient is $>$ 0.875 and the percentage of accepted sequences is between [85; 100] 
    \end{itemize}

For ERG context : 
    \begin{itemize}
        \item for k $>$ 50, average value of silhouette coefficient is $>$ 0.750 and the percentage of accepted sequences is between [85;100 ] 
        \item for k $>$ 100, average value of silhouette coefficient is $>$ 0.825 and the percentage of accepted sequences is between [87;100 ] 
    \end{itemize}

These results show that it is possible, using the hidden representation of the hidden layer of a RNN-LSTM, to extract a knowledge representation of the hidden rules that is close to the original grammar by 80\% in the worst case. 
On top of that, the analysis of our results shows that our algorithm converges. In the case of RG and ERG, an increase in k makes it possible to extract automata that recognize a larger number of sequences.
Let us underline the importance of this result because it implies that the choice of the size of k is not a limitation of our algorithm, but a compromise to be achieved according to the precision sought. Indeed, to have more precise results, it is enough to increase k (the number of clusters), at the cost of a greater need for computing resources.
The results contribute to interpret the k value in the k-means clustering as an adjusting variable : when k increases, it induces more precise knowledge representation but we loose the generality of the representation. In other words, what is earned in accuracy is lost in generality.
k is therefore a cursor to be adjusted according to the level of interpretability we want for a situation.
This leads us to another important remark : the methodology we propose here provides flexible results depending on the value of k and the available computing power.


\subsection{Real data set (electrical diagrams)}

To test the applicability of our approach on real data, we apply our approach on sequences generated from an electrical grammar. Indeed, previous works shed light on the existence of hidden knowledge inside sequences of electrical components \citep{chraibikaadoud:hal-01525028}. In this specific case, the grammar used to generate the sequences was designed and validated by experts from the electrical field. This grammar is composed of 25 symbols corresponding to 25 different electrical components. We followed the same procedure as before: a learning phase on 200 000 valid sequences and then 10 tests simulations on 20 000 valid sequences.

\begin{figure}[htbp]
\centering
\includegraphics[width=\textwidth]{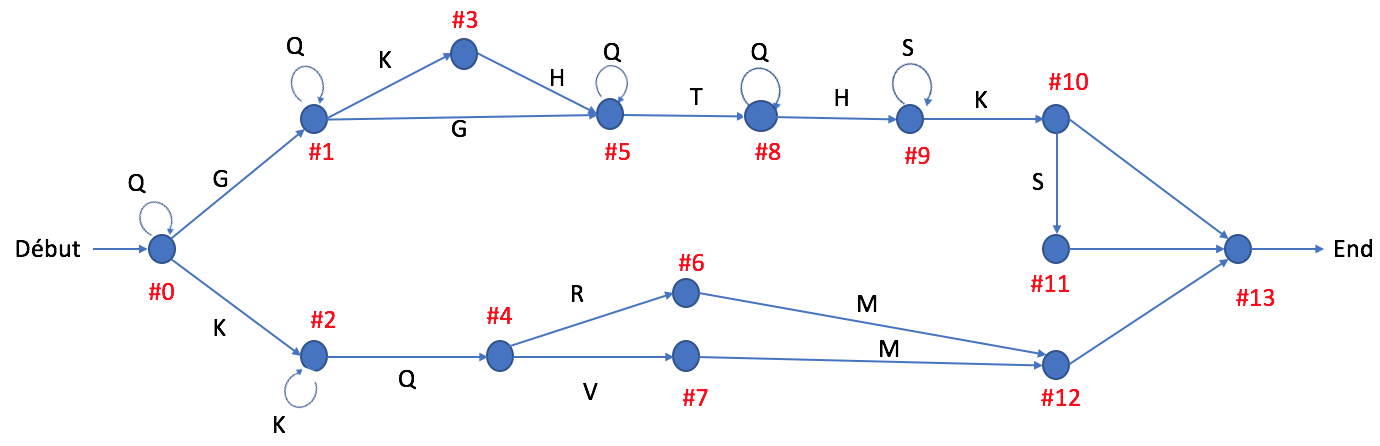}
\caption{\label{fig:grammaireEL} Electrical grammar from the study of 3 electrical diagrams}
\end{figure}

\begin{figure}[htbp]
\centering
\includegraphics[width=\textwidth]{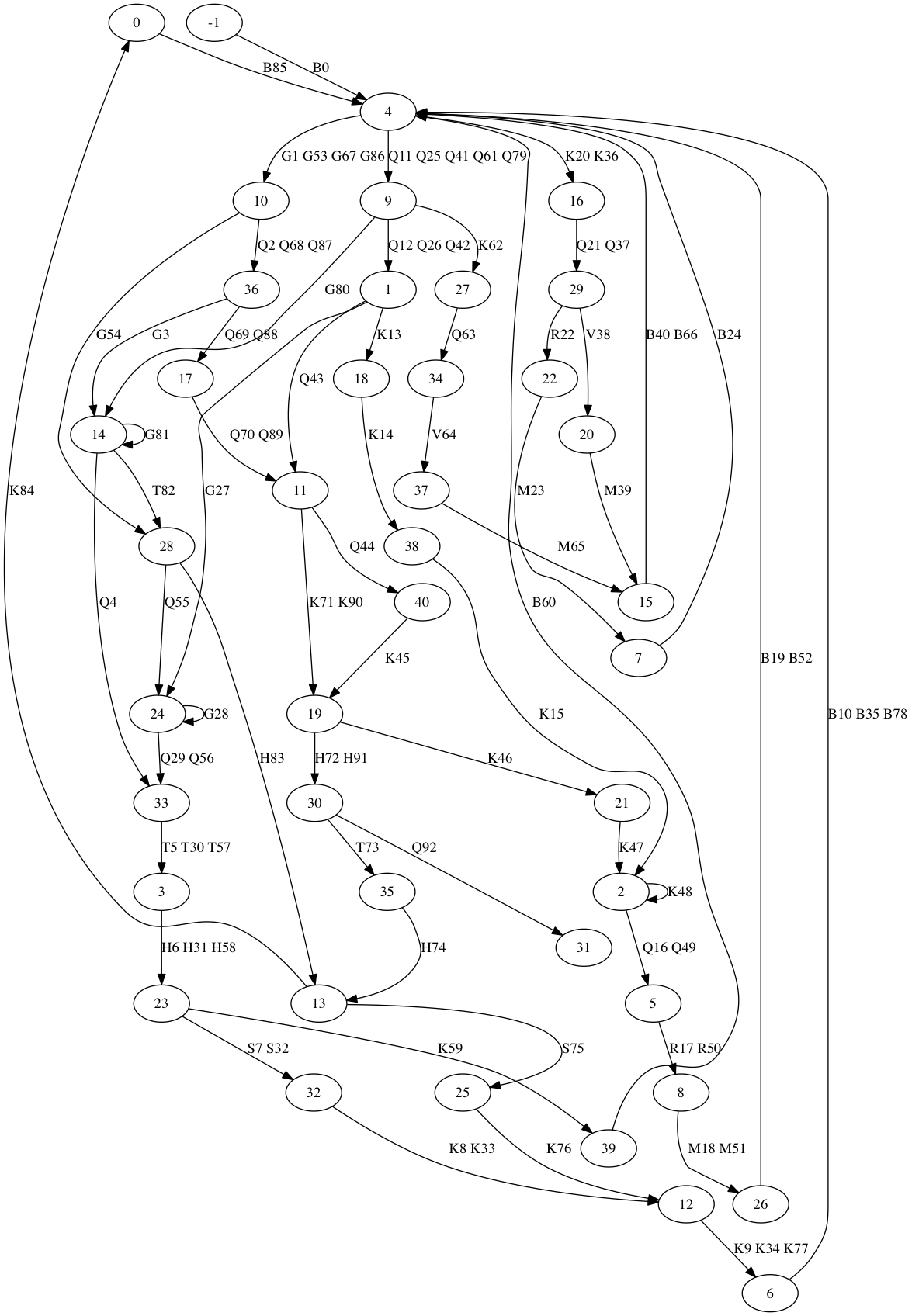}
\caption{\label{fig:EL_KMeans_Matrix_DFA2_41}A long-label FSA  obtained with a k-means algorithm for clustering, where k=41 and from an RNN-LSTM having learned sequences of electrical components. Extraction was performed on the first 79 time steps.
}
\end{figure}

The extracted FSA in this specific context was validated by being submitted to experts in the electrical field. In this specific context we have privileged human validation rather than automatic one, since the electrical grammar was hand made and that we were mostly interested in the potential discrepancies between extracted electrical rules (automata) with the knowledge of experts in the domain. Even though these are preliminary results on real data, they have shown that this approach can be transposed to actual data.

\section{Discussion} \label{section_Discussion}

Through the presented work, two main achievements were made : First, modeling implicit learning using LSTM, and second defining a methodology to extract implicit rules in an interpretable shape from RNN-LSTM that learn sequences. 

Considering the first achievement, we manage to do with LSTM what failed before with SRN : Indeed, several attempts have been made, particularly with SRN to extract a stable and implicit representation of encoded rules hidden in sequences, but they failed because the SRN does not manage ambiguous sequences, with long sequential dependencies, i.e. increasingly complex data.
On the contrary, the LSTMs were able to learn and generate sequences of variable size and characterized by the existence of strong sequential dependencies generated from non-binary artificial grammars. This ability is due to the unique characteristics of LSTM models, including short-term memory (activity patterns at each time step), long-term memory (weights), and intermediate memory (CEC of LSTM units).

The second achievement is about interpretability : 
We assumed that by implicit learning, 1) important knowledge about grammatical rules corresponding to the valid transitions and their context in sequences was encoded in the hidden layer of the RNN-LSTM and 2) that it could be extracted and explicitly represented as an automaton.
By only processing the output activity patterns of the hidden layer, we were able to extract a representation in the form of graphs, with different notation systems (cf figures \ref{fig:RulesExtracted3graphes}, \ref{fig:GramRG_graphRG}, \ref{fig:GramERG_graphERG}), each carrying information on the internal functioning of the RNN-LSTM.
The representation without notation informs about the arrangement of states and transitions between them, the notation with long labels informs about the temporal arrangement of patterns between the different states, and offers a contextual explanation regarding the management of patterns by the RNN-LSTM.
Finally, the representation with simplified notation provides a synthetic and explicit representation of the grammatical rules learned and governing the predictions of the RNN-LSTM, to be directly compared with the original grammars.
Figure \ref{fig:GramRG_graphRG} shows a comparison between the Reber grammar and the extracted automaton with simplified notation, and figure  \ref{fig:GramERG_graphERG} in an ERG context. 
The final automaton with simplified labels, once minimized, has been fed with valid sequences for each considered grammar.
Over 10 consecutive simulations, the percentage of recognition of valid sequences is above 80\% for k $>50$ for RG and ERG. 
Finally, we have shown that this percentage is not a limit in itself of our algorithm, but a compromise to be made between the degree of precision desired during the extraction process and the computing power allocated.
The k value in the k-means algorithm is thus a important parameter for the interpretability : according to its value and the expected interpretability, it permits to extract an automaton that can explain the local behavior (for a specific prediction) or the global behavior. 
It is thus not the accuracy of the extracted automaton that it important but its relevance regarding the initial question of the black box RNN.
The most difficult point is thus to determine "what is a good representation?", which is a context and a human dependant question.

At the application level, being able to cope with abstract but rather complex grammars, we showed through preliminary results that our approach could be transposed to more concrete fields. 
We also stress the fact that, even if grammars were known here, this knowledge was not used for automaton extraction and the process could consequently be applied to unknown grammars, as it is often the case in concrete applications.
In the industrial domain, the work made on sequences of electrical components where the grammar is only known by experts, permitted to formalize an electrical grammar, that allowed to generate sequences of electrical components for the RNN-LSTM to learn. After applying a part of the extraction process defined in this paper, the first results submitted to experts specialized in drawing electrical diagrams showed that the extracted rules hidden in those sequences did indeed contain electrical connectivity rules but also habits of the engineers that draw electrical diagrams \citep{chraibikaadoud:hal-01525028, chraibikaadoud:tel-01771685}. 
This makes us propose that, more generally, this process of knowledge extraction could be interesting to study expertise in many fields, which represents a very precious knowledge in many companies but is most of the time implicit and consequently very difficult to transfer to other people.

More globally, all this work indeed addresses the problem of interpretability in ANNs seen as black boxes and particularly about RNNs, which are particularly difficult to understand as they process sequences and not just patterns. We have just evoked that our techniques can provide valuable information at different levels of details, for local as well as global interpretability. When the learning has been done, our algorithms allow to consider any new sequence and to explain in details how it is processed by the RNN in its prediction task. 
An important perspective of this work would be to work on a richer information about the hidden space than the activity of the output of the hidden layer. In particular, to explain how LSTMs support sequential dependencies, it would be interesting to explore the hidden representation at the level of activity patterns of LSTM gates, cells, as well as CECs.
Note that this question also arises if we apply our approach to variations of standard LSTMs, such as Gated Recurrent Units (GRU) or LSTMs with peepholes connections. A network with additional hidden layers could also be considered to study the abstraction of the implicit representation encoded by the network from one layer to another.
All these characteristics and perspectives may thus participate to improve trust in machine learning algorithms by making them more accessible to as many people as possible.\\

\section{Acknowledgement} \label{section_acknowledgement}

This work was initiated as part of a thesis in collaboration with Inria and the Algo'Tech company (Bidart, France), specialized in computer design and drawing software in the field of electrical diagrams. 
We would like to thank all the actors who made this work possible.

\clearpage

\renewcommand*{\bibfont}{\footnotesize}
\bibliography{biblio}

\begin{thebibliography}{10}

\bibitem{abu90}
Y.~Abu-Mostafa.
\newblock Learning from hints in neural networks.
\newblock {\em Journal of Complexity}, 6:192--198, 1990.

\bibitem{ayache:hal-01888514}
St{\'e}phane Ayache, R{\'e}mi Eyraud, and No{\'e} Goudian.
\newblock {Explaining Black Boxes on Sequential Data using Weighted Automata}.
\newblock In {\em {14th International Conference on Grammatical Inference}},
  volume~88, Wroclaw,, Poland, September 2018.

\bibitem{bengio2013representation}
Yoshua Bengio, Aaron Courville, and Pascal Vincent.
\newblock Representation learning: A review and new perspectives.
\newblock {\em IEEE transactions on pattern analysis and machine intelligence},
  35(8):1798--1828, 2013.

\bibitem{blanco2000extracting}
Armando Blanco, Miguel Delgado, and MC~Pegalajar.
\newblock Extracting rules from a (fuzzy/crisp) recurrent neural network using
  a self-organizing map.
\newblock {\em International Journal of Intelligent Systems}, 15(7):595--621,
  2000.

\bibitem{chraibikaadoud:tel-01771685}
Ikram Chraibi~Kaadoud.
\newblock {\em {Sequence learning and rules extraction from recurrent neural
  networks : application to the drawing of technical diagrams}}.
\newblock Theses, {Universit{\'e} de Bordeaux}, March 2018.

\bibitem{chraibikaadoud:hal-01525028}
Ikram Chraibi~Kaadoud, Nicolas~P. Rougier, and Fr{\'e}d{\'e}ric Alexandre.
\newblock {Implicit knowledge extraction and structuration from electrical
  diagrams}.
\newblock In {\em {The 30th International Conference on Industrial,
  Engineering, Other Applications of Applied Intelligent Systems}}, Arras,
  France, June 2017.

\bibitem{cleeremans1993mechanisms}
Axel Cleeremans.
\newblock {\em Mechanisms of implicit learning: Connectionist models of
  sequence processing}.
\newblock MIT press, 1993.

\bibitem{cleeremans1991learning}
Axel Cleeremans and James~L McClelland.
\newblock Learning the structure of event sequences.
\newblock {\em Journal of Experimental Psychology: General}, 120(3):235, 1991.

\bibitem{cleeremans1989finite}
Axel Cleeremans, David Servan-Schreiber, and James~L McClelland.
\newblock Finite state automata and simple recurrent networks.
\newblock {\em Neural computation}, 1(3):372--381, 1989.

\bibitem{elman1990finding}
Jeffrey~L Elman.
\newblock Finding structure in time.
\newblock {\em Cognitive science}, 14(2):179--211, 1990.

\bibitem{gasparini2004implicit}
Silvia Gasparini.
\newblock Implicit versus explicit learning: Some implications for l2 teaching.
\newblock {\em European Journal of Psychology of Education}, 19(2):203--219,
  2004.

\bibitem{gers1999learning}
Felix~A Gers, J{\"u}rgen Schmidhuber, and Fred Cummins.
\newblock Learning to forget: Continual prediction with lstm.
\newblock 1999.

\bibitem{giles1991second}
C~Lee Giles, D~Chen, CB~Miller, HH~Chen, GZ~Sun, and YC~Lee.
\newblock Second-order recurrent neural networks for grammatical inference.
\newblock In {\em Neural Networks, 1991., IJCNN-91-Seattle International Joint
  Conference on}, volume~2, pages 273--281. IEEE, 1991.

\bibitem{giles1992learning}
C~Lee Giles, Clifford~B Miller, Dong Chen, Hsing-Hen Chen, Guo-Zheng Sun, and
  Yee-Chun Lee.
\newblock Learning and extracting finite state automata with second-order
  recurrent neural networks.
\newblock {\em Neural Computation}, 4(3):393--405, 1992.

\bibitem{gombert2006epi}
Jean~Emile Gombert.
\newblock Epi/m{\'e}ta vs implicite/explicite: niveau de contr{\^o}le cognitif
  sur les traitements et apprentissage de la lecture.
\newblock {\em Langage \& pratiques}, 38:68--76, 2006.

\bibitem{Goodfellow2016Convolutional}
Ian Goodfellow, Yoshua Bengio, and Aaron Courville.
\newblock Convolutional networks.
\newblock In {\em Deep Learning}, chapter~9, pages 330--372. MIT Press, 2016.

\bibitem{greff2017lstm}
Klaus Greff, Rupesh~K Srivastava, Jan Koutn{\'\i}k, Bas~R Steunebrink, and
  J{\"u}rgen Schmidhuber.
\newblock Lstm: A search space odyssey.
\newblock {\em IEEE transactions on neural networks and learning systems},
  2017.

\bibitem{DBLP:journals/corr/abs-1802-01933}
Riccardo Guidotti, Anna Monreale, Franco Turini, Dino Pedreschi, and Fosca
  Giannotti.
\newblock A survey of methods for explaining black box models.
\newblock {\em CoRR}, abs/1802.01933, 2018.

\bibitem{hochreiter1997long}
Sepp Hochreiter and J{\"u}rgen Schmidhuber.
\newblock Long short-term memory.
\newblock {\em Neural computation}, 9(8):1735--1780, 1997.

\bibitem{Hunter:2007}
J.~D. Hunter.
\newblock Matplotlib: A 2d graphics environment.
\newblock {\em Computing In Science \& Engineering}, 9(3):90--95, 2007.

\bibitem{jacobsson2005rule}
Henrik Jacobsson.
\newblock Rule extraction from recurrent neural networks: Ataxonomy and review.
\newblock {\em Neural Computation}, 17(6):1223--1263, 2005.

\bibitem{Jones:2001}
Eric Jones, Travis Oliphant, and Pearu Peterson.
\newblock {SciPy}: Open source scientific tools for {Python}, 2001.

\bibitem{keller2001classifiers}
Robert~M Keller.
\newblock Classifiers, acceptors, transducers, and sequencers.
\newblock {\em Computer Science: Abstraction to Implementation. Harvey Mudd
  College}, page 480, 2001.

\bibitem{lapalme2006composition}
Jasmin Lapalme.
\newblock Composition automatique de musique {\`a} l'aide de r{\'e}seaux de
  neurones r{\'e}currents et de la structure m{\'e}trique.
\newblock 2006.

\bibitem{lipton2016}
Zachary Lipton.
\newblock The mythos of model interpretability.
\newblock In {\em Proceedings of the ICML Workshop on Human Interpretability in
  Machine Learning}, 2016.

\bibitem{nadeau2011connaissances}
Marie Nadeau and Carole Fisher.
\newblock Les connaissances implicites et explicites en grammaire: quelle
  importance pour l'enseignement? quelles cons{\'e}quences?
\newblock {\em Bellaterra journal of teaching and learning language and
  literature}, 4(4):0001--31, 2011.

\bibitem{nerode1958linear}
Anil Nerode.
\newblock Linear automaton transformations.
\newblock {\em Proceedings of the American Mathematical Society},
  9(4):541--544, 1958.

\bibitem{omlin1996extraction}
Christian~W Omlin and C~Lee Giles.
\newblock Extraction of rules from discrete-time recurrent neural networks.
\newblock {\em Neural networks}, 9(1):41--52, 1996.

\bibitem{oudeyer2009auto}
Pierre-Yves Oudeyer.
\newblock L'auto-organisation dans l'{\'e}volution de la parole, 2009.

\bibitem{pascual1995procedural}
Alvaro Pascual-Leone, Jordan Grafman, and Mark Hallett.
\newblock Procedural learning and prefrontal cortex.
\newblock {\em Annals of the New York Academy of Sciences}, 769(1):61--70,
  1995.

\bibitem{scikit-learn}
F.~Pedregosa, G.~Varoquaux, A.~Gramfort, V.~Michel, B.~Thirion, O.~Grisel,
  M.~Blondel, P.~Prettenhofer, R.~Weiss, V.~Dubourg, J.~Vanderplas, A.~Passos,
  D.~Cournapeau, M.~Brucher, M.~Perrot, and E.~Duchesnay.
\newblock Scikit-learn: Machine learning in {P}ython.
\newblock {\em Journal of Machine Learning Research}, 12:2825--2830, 2011.

\bibitem{perruchet_implicit_2006}
Pierre Perruchet and Sebastien Pacton.
\newblock Implicit learning and statistical learning: one phenomenon, two
  approaches.
\newblock {\em Trends in Cognitive Sciences}, 10(5):233--238, May 2006.

\bibitem{pothos_theories_2007}
Emmanuel~M. Pothos.
\newblock Theories of artificial grammar learning.
\newblock {\em Psychological Bulletin}, 133(2):227--244, 2007.

\bibitem{reber1967implicit}
Arthur~S Reber.
\newblock Implicit learning of artificial grammars.
\newblock {\em Journal of verbal learning and verbal behavior}, 6(6):855--863,
  1967.

\bibitem{reber_neural_2013}
Paul~J. Reber.
\newblock The neural basis of implicit learning and memory: {A} review of
  neuropsychological and neuroimaging research.
\newblock {\em Neuropsychologia}, 51(10):2026--2042, August 2013.

\bibitem{remm02}
J.-F. Remm and F.~Alexandre.
\newblock Knowledge extraction using artificial neural networks : Application
  to radar target identification.
\newblock {\em Signal Processing}, 82(1), 2002.

\bibitem{rousseeuw1987silhouettes}
Peter~J Rousseeuw.
\newblock Silhouettes: a graphical aid to the interpretation and validation of
  cluster analysis.
\newblock {\em Journal of computational and applied mathematics}, 20:53--65,
  1987.

\bibitem{rumel86}
D.~Rumelhart and J.~MacClelland, editors.
\newblock {\em {Parallel distributed processing}}.
\newblock MIT Press, Cambridge, 1986.

\bibitem{schellhammer1998knowledge}
Ingo Schellhammer, Joachim Diederich, Michael Towsey, and Claudia Brugman.
\newblock Knowledge extraction and recurrent neural networks: An analysis of an
  elman network trained on a natural language learning task.
\newblock In {\em Proceedings of the joint conferences on new methods in
  language processing and computational natural language learning}, pages
  73--78. Association for Computational Linguistics, 1998.

\bibitem{ServanSchreiber1988encoding}
D~Servan-Schreiber, A~Cleermans, and JL~McClelland.
\newblock Encoding sequential structure in simple recurrent networks.
\newblock {\em School of Computer Science, Carnegie-Mellon University,
  Pittsburgh, PA, CMU-CS-88-183.(November, 1988)}, 1988.

\bibitem{servan1989learning}
David Servan-Schreiber, Axel Cleeremans, and James~L McClelland.
\newblock Learning sequential structure in simple recurrent networks.
\newblock In {\em Advances in neural information processing systems}, pages
  643--652, 1989.

\bibitem{servan1991graded}
David Servan-Schreiber, Axel Cleeremans, and James~L McClelland.
\newblock Graded state machines: The representation of temporal contingencies
  in simple recurrent networks.
\newblock {\em Machine Learning}, 7(2):161--193, 1991.

\bibitem{setiono96}
R.~Setiono and H.~Liu.
\newblock Symbolic representation of neural networks.
\newblock {\em Computer}, 29(3):71--77, 1996.

\bibitem{tivno1995learning}
Peter Ti{\v{n}}o and Jozef {\v{S}}ajda.
\newblock Learning and extracting initial mealy automata with a modular neural
  network model.
\newblock {\em Neural Computation}, 7(4):822--844, 1995.

\bibitem{tomita1982dynamic}
Masaru Tomita.
\newblock Dynamic construction of finite-state automata from examples using
  hill-climbing.
\newblock In {\em Proceedings of the Fourth Annual Conference of the Cognitive
  Science Society}, pages 105--108, 1982.

\bibitem{Walt:2011}
Stéfan van~der Walt, S~Chris Colbert, and Gaël Varoquaux.
\newblock The {NumPy} array: A structure for efficient numerical computation.
\newblock {\em Computing in Science {\&} Engineering}, 13(2):22--30, 3 2011.

\bibitem{wang2017empirical}
Qinglong Wang, Kaixuan Zhang, Alexander G~Ororbia II, Xinyu Xing, Xue Liu, and
  C~Lee Giles.
\newblock An empirical evaluation of rule extraction from recurrent neural
  networks.
\newblock {\em arXiv preprint arXiv:1709.10380}, 2017.

\bibitem{wang2018comparison}
Qinglong Wang, Kaixuan Zhang, II~Ororbia, G~Alexander, Xinyu Xing, Xue Liu, and
  C~Lee Giles.
\newblock A comparison of rule extraction for different recurrent neural
  network models and grammatical complexity.
\newblock {\em arXiv preprint arXiv:1801.05420}, 2018.

\bibitem{watrous1992induction}
Raymond~L Watrous and Gary~M Kuhn.
\newblock Induction of finite-state automata using second-order recurrent
  networks.
\newblock In {\em Advances in neural information processing systems}, pages
  309--317, 1992.

\bibitem{DBLP:journals/corr/abs-1711-09576}
Gail Weiss, Yoav Goldberg, and Eran Yahav.
\newblock Extracting automata from recurrent neural networks using queries and
  counterexamples.
\newblock {\em CoRR}, abs/1711.09576, 2017.

\bibitem{weiss2017extracting}
Gail Weiss, Yoav Goldberg, and Eran Yahav.
\newblock Extracting automata from recurrent neural networks using queries and
  counterexamples.
\newblock {\em arXiv preprint arXiv:1711.09576}, 2017.

\bibitem{zeng1993learning}
Zheng Zeng, Rodney~M Goodman, and Padhraic Smyth.
\newblock Learning finite state machines with self-clustering recurrent
  networks.
\newblock {\em Learning}, 5(6), 1993.

\end{thebibliography}
\bibliographystyle{plain}

\end{document}